\documentclass[sigconf,nonacm]{acmart} 

\usepackage{booktabs} 
\usepackage{enumitem}
\usepackage{diagbox}
\usepackage{amsmath}
\usepackage{xspace}
\usepackage{soul}
\usepackage{fancyvrb}

\usepackage[ruled]{algorithm2e} 

\SetAlFnt{\small}
\SetAlCapFnt{\small}
\SetAlCapNameFnt{\small}
\SetAlCapHSkip{0pt}

\newcommand{\sysname}{Direct-a-Video\xspace}
\newcommand{\R}[1]{{\color{black}#1}} 
\definecolor{softgreen}{RGB}{153, 204, 153}
\newcommand{\A}[1]{#1} 

\copyrightyear{2024} 
\acmYear{2024} 
\setcopyright{acmlicensed}\acmConference[SIGGRAPH Conference Papers '24]{Special Interest Group on Computer Graphics and Interactive Techniques Conference Conference Papers '24}{July 27-August 1, 2024}{Denver, CO, USA}
\acmBooktitle{Special Interest Group on Computer Graphics and Interactive Techniques Conference Conference Papers '24 (SIGGRAPH Conference Papers '24), July 27-August 1, 2024, Denver, CO, USA}
\acmDOI{10.1145/3641519.3657481}
\acmISBN{979-8-4007-0525-0/24/07}

\begin{document}
\title{Direct-a-Video: Customized Video Generation with User-Directed Camera Movement and Object Motion}

\author{Shiyuan Yang}
\affiliation{%
  \institution{City University of Hong Kong$^1$}  
  \institution{Tianjin University$^2$}  
  \city{$^1$Hong Kong, $^2$Tianjin}
  \country{China}
}
  \email{s.y.yang@my.cityu.edu.hk}

\author{Liang Hou}
\affiliation{%
  \institution{Kuaishou Technology}
  \city{Beijing}
  \country{China}}
    \email{houliang06@kuaishou.com}

\author{Haibin Huang}
\affiliation{%
  \institution{Kuaishou Technology}
  \city{Beijing}
  \country{China}}
    \email{jackiehuanghaibin@gmail.com}

\author{Chongyang Ma}
\affiliation{%
  \institution{Kuaishou Technology}
  \city{Beijing}
  \country{China}}
    \email{chongyangm@gmail.com}

\author{Pengfei Wan}
\affiliation{%
  \institution{Kuaishou Technology}
  \city{Beijing}
  \country{China}}
\email{wanpengfei@kuaishou.com}

\author{Di Zhang}
\affiliation{%
  \institution{Kuaishou Technology}
  \city{Beijing}
  \country{China}}
\email{zhangdi08@kuaishou.com}

\author{Xiaodong Chen}
\affiliation{%
  \institution{Tianjin University}
  \city{Tianjin}
  \country{China}}
\email{xdchen@tju.edu.cn}

\author{Jing Liao}
\authornote{Corresponding author.}
\affiliation{%
\institution{City University of Hong Kong}
  \city{Hong Kong}
  \country{China}}
\email{jingliao@cityu.edu.hk}

\renewcommand{\shorttitle}{Direct-a-Video}
\renewcommand{\shortauthors}{Yang, S. et al}

\begin{abstract}
Recent text-to-video diffusion models have achieved impressive progress. In practice, users often desire the ability to control object motion and camera movement independently for customized video creation. However, current methods lack the focus on separately controlling object motion and camera movement in a decoupled manner, which limits the controllability and flexibility of text-to-video models. In this paper, we introduce Direct-a-Video, a system that allows users to independently specify motions for multiple objects as well as camera's pan and zoom movements, as if directing a video. We propose a simple yet effective strategy for the decoupled control of object motion and camera movement. Object motion is controlled through \textit{spatial cross-attention} modulation using the model's inherent priors, requiring no additional optimization. For camera movement, we introduce new \textit{temporal cross-attention} layers to interpret quantitative camera movement parameters. We further employ an augmentation-based approach to train these layers in a self-supervised manner on a small-scale dataset, eliminating the need for explicit motion annotation. Both components operate independently, allowing individual or combined control, and can generalize to open-domain scenarios. Extensive experiments demonstrate the superiority and effectiveness of our method. Project page and code are available at 
\url{https://direct-a-video.github.io/}.
\end{abstract}


\graphicspath{
{figs/}
}

%
%

\newcommand{\etal}{\emph{et al.}}
\newcommand{\ie}{\emph{i.e.}}
\newcommand{\eg}{\emph{e.g.}}

\begin{teaserfigure}
  \includegraphics[width=\textwidth]{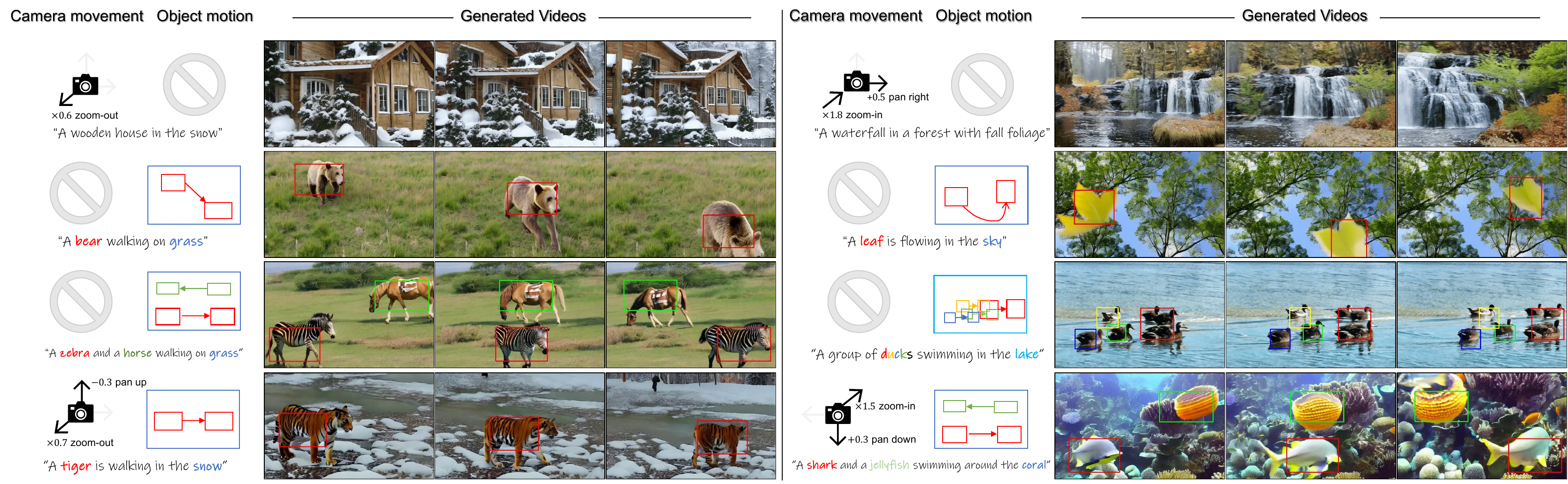}
  \caption{Direct-a-Video is a text-to-video generation framework that allows users to individually or jointly control the camera movement and/or object motion.}
  \label{fig.teaser}
\end{teaserfigure}

\maketitle

\section{Introduction} \label{sec.intro}

Text-to-image (T2I) diffusion models have already demonstrated astonishingly high quality and diversity in image generation and editing~\cite{ddpm, dalle2, ldm, imagen, glide, unipaint, luckydog1, luckydog2}. The rapid development of T2I diffusion models has also spurred the recent emergence of text-to-video (T2V) diffusion models~\cite{videoldm, imagenvideo, makeavideo, modelscope, svd}, which are normally extended from pretrained T2I models for video generation and editing. On the other hand, the advent of controllable techniques in T2I models, such as ControlNet~\cite{controlnet}, T2I-adapter~\cite{t2iadapter} and GLIGEN~\cite{gligen}, has allowed users to specify the spatial layout of generated images through conditions like sketch maps, depth maps, or bounding boxes etc., significantly enhancing the spatial controllability of T2I models. Such spatial controllable techniques have also been successfully extended to spatial-temporal control for video generation. One of the representative works in this area is VideoComposer~\cite{videocomposer}, which can synthesize a video given a sequence of sketch or motion vector maps.

Despite the success of video synthesis, current T2V methods often lack support for user-defined and disentangled control over camera movement and object motion, which limits the flexibility in video motion control. In a video, both objects and the camera exhibit their respective motions. Object motion originates from the subject's activity, while camera movement influences the transition between frames. The overall video motion becomes well-defined only when both camera movement and object motion are determined. For example, focusing solely on object motion, such as generating a video clip where an object moves to the right within the frame, can lead to multiple scenarios. The camera may remain stationary while the object itself moves right, or the object may be stationary while the camera moves left, or both the object and the camera may be moving at different speeds. This ambiguity in the overall video motion can arise. Therefore, the decoupling and independent control of camera movement and object motion not only provide more flexibility but also reduce ambiguity in the video generation process. However, this aspect has received limited research attention thus far.

To control camera movement and object motion in T2V generation, a straightforward approach would be to follow the supervised training route similar to works like VideoComposer~\cite{videocomposer}. Following such kind of scheme involves training a conditional T2V model using videos annotated with both camera and object motion information. However, this would bring the following challenges:
(1) In many video clips, object motion is often coupled with camera movements due to their inherent correlation. For example, when a foreground object moves to some direction, the camera typically pans in the same direction due to the preference to keep the main subject at the center of the frame. Training on such coupled camera and object motion data makes it difficult for the model to distinguish between camera movements and and object motion.
(2) Obtaining large-scale video datasets with complete camera movement and object motion annotations is challenging due to the laborious and costly nature of performing frame-by-frame object tracking and camera pose estimation. Additionally, training a video model on a large-scale dataset can be computationally expensive.

In this work, we introduce \sysname, a text-to-video framework that enables users to independently specify the camera's pan and zoom movements and the motions of scene objects, allowing them to create their desired motion pattern as if they were directing a video (Figure~\ref{fig.teaser}).
To achieve this, we propose a strategy for decoupling camera and object control by employing two orthogonal controlling mechanisms. In essence, we learn the camera movement through a self-supervised and lightweight training approach. Conversely, during inference, we adopt a training-free method to control object motion. Our strategy avoids the need for intensive collection of motion annotations and video grounding datasets.

In camera movement control, we train an additional module to learn the frame transitions. Specifically, we introduce new temporal cross-attention layers, known as the camera module, which functions similarly to spatial cross-attention in interpreting textual language.
This camera module interprets ``camera language'', specifically camera panning and zooming parameters, enabling precise control over camera movement. However, acquiring datasets with camera movement annotations can pose a challenge. To overcome this laborious task, we employ a self-supervised training strategy that relies on camera movement augmentation. This approach eliminates the need for explicit motion annotations. Importantly, we train these new layers while preserving the original model weights, ensuring that the extensive prior knowledge embedded within the T2V model remains intact. Although the model is initially trained on a small-scale video dataset, it acquires the capability to quantitatively control camera movement in diverse, open-domain scenarios.

In object motion control, a significant challenge arises from the availability of well-annotated grounding datasets for videos, curating such datasets is often a labor-intensive process. To bypass these issues, we draw inspiration from previous attention-based image-layout control techniques in T2I models~\cite{p2p, densediff}. We utilize the internal priors of the T2V model through spatial cross-attention modulation, which is a training-free approach, thereby eliminating the need for collecting grounding datasets and annotations for object motion. To facilitate user interaction, we enable users to specify the spatial-temporal trajectories of objects by drawing bounding boxes at the first and last frames, as well as the intermediate path. Such interaction is simpler and more user-friendly compared to previous pixel-wise control methods~\cite{videocomposer}.

Given that our approach independently controls camera movement and object motion, thereby effectively decouples the two, offering users enhanced flexibility to individually or simultaneously manipulate these aspects in video creation. 

In summary, our contributions are as follows:
\begin{itemize}[leftmargin=*]
\item We propose a unified framework for controllable video generation that decouples camera movement and object motion, allowing users to independently or jointly control both aspects.

\item For camera movement, we introduce a novel temporal cross-attention module dedicated to camera movement conditioning. This camera module is trained through self-supervision, enabling users to quantitatively specify the camera's horizontal and vertical panning speeds, as well as its zooming ratio.

\item For object motion, we utilize a training-free spatial cross-attention modulation, enabling users to easily define the motion trajectories for one or more objects by drawing bounding boxes.

\end{itemize}

\section{Related Work} \label{sec.rw}
\subsection{Text-to-Video Generation}
\label{sec.rw.t2v}
The success of text-to-image (T2I) models has revealed their potential for text-to-video (T2V) generation. T2V models are often evolved from T2I models by incorporating temporal layers. Early T2V models \cite{vdm, imagenvideo, makeavideo} perform the diffusion process in pixel space, which requires multiple cascaded models to generate high-resolution or longer videos, resulting in high computational complexity. Recent T2V models draw inspiration from latent diffusion \cite{ldm} and operate in a lower-dimensional and more compact latent space \cite{videoldm, magicvideo, gen1, modelscope, animatediff}. The most recent Stable Video Diffusion \cite{svd} utilizes curated training data and is capable of generating high-quality videos.

On the other hand, the development of T2I editing techniques \cite{p2p,nulltxtinv,txtinv,db,customdiff} has facilitated zero/few-shot video editing tasks. These techniques convert a given source video to a target video through approaches such as weight fine-tuning \cite{tav}, dense map conditioning \cite{renderavideo,gen1,controlvideo, tokenflow}, sparse point conditioning \cite{videoswap, dift}, attention feature editing \cite{fatezero,videop2p,pix2video,v2vzero}, and canonical space processing \cite{codef,stablevideo, lna}. Some works specifically focus on synthesizing human dance videos using source skeleton sequences and reference portraits \cite{magicdance, animateanyone, magicanimate, dreamoving, disco}, which have yielded impressive results.

\subsection{Video Generation with Controllable Motion}
\label{sec.rw.motion}
As motion is an important factor in video, research on video generation with motion control has garnered increasing attention. We can categorize the works in this field into three groups based on the type of input media: image-to-video, video-to-video, and text-to-video.

\paragraph{Image-to-video} Some methods focus on transforming static images into videos, and a popular approach for motion control is through key point dragging \cite{dragnvwa, mcdiff, dragvideo}. While this interaction method is intuitive and user-friendly, it has limitations due to the local and sparse nature of the key points. Consequently, its capacity for controlling motion at a large granularity is significantly restricted.

\paragraph{Video-to-video} These works primarily focus on motion transfer, which involves learning a specific subject action from source videos and applying it to target videos using various techniques, including fine-tuning the model on a set of reference videos with similar motion patterns \cite{lamp, motiondirector, vmc, dreamvideo}, or borrowing spatial features (\eg, sketch, depth maps) \cite{videocomposer, controlavideo} or sparse features (\eg, DIFT point embedding) \cite{videoswap} from source videos. These methods highly rely on the motion priors from the source videos, which, however, are not always practically available.

\paragraph{Text-to-video} In the case where the source video is unavailable, generating videos from text with controllable motion is a meaningful but relatively less explored task. Our work focuses on this category. Existing approaches in this category include AnimateDiff \cite{animatediff}, which utilizes ad-hoc motion LoRA modules \cite{lora} to enable specific camera movements. However, it lacks quantitative camera control and also does not support object motion control. VideoComposer ~\cite{videocomposer} provides global motion guidance by conditioning on pixel-wise motion vectors. However, the dense control manner offered by VideoComposer is inefficient to use and does not explicitly separate camera and object motion, resulting in inconvenient user interaction. A concurrent work, Peekaboo \cite{peekaboo}, also uses bounding boxes to control the trajectory of the object through attention masking. However, their method originally does not consider multi-object scenarios and also does not support control over camera movement, unlike our approach.
MotionCtrl \cite{motionctrl}, another concurrent work, allows for separate 2D point-driven object motion control and 3D trajectory-driven camera control by training camera and object control modules. However, its training preparation is labor-intensive, requiring the extraction of motion trajectories on large-scale video dataset. Moreover, it struggles to control multiple different objects with varied motion directions, as it lacks the explicit binding between objects and their motion trajectories during the training.
In contrast, our self-supervised training scheme does not require any motion annotations and can achieve motion control over camera and multiple objects, bringing more flexibility for video synthesis.

\section{Method} \label{sec.meth}
\begin{figure*}[t]
\begin{center}
   \includegraphics[width=\linewidth]{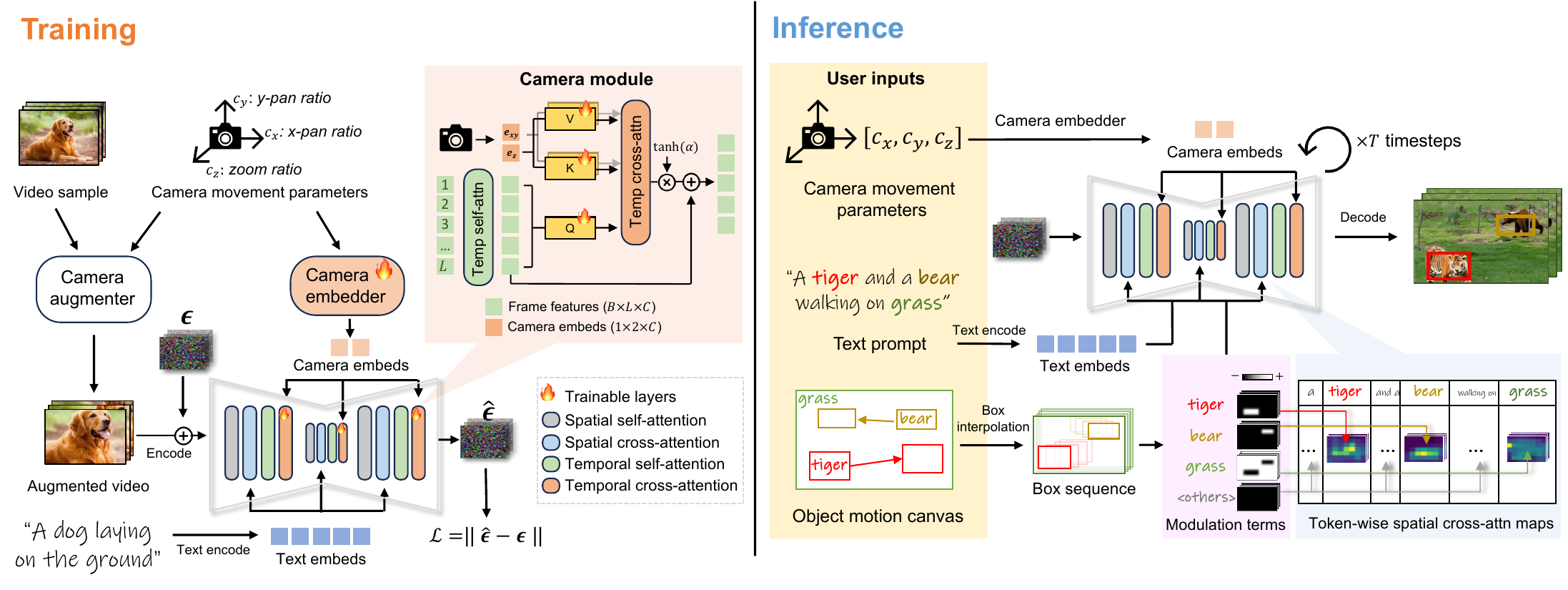}
\end{center}
\caption{The overall pipeline of \sysname.
The camera movement is learned in the training stage and the object motion is implemented in the inference stage.  \textbf{Left}: During training, we apply augmentation to video samples to simulate camera movement using panning and zooming parameters. These parameters are embedded and injected into newly introduced temporal cross-attention layers as the camera movement  conditioning, eliminating the need for camera movement annotation.
\textbf{Right}: During inference, along with camera movement, user inputs a text prompt containing object words and associated box trajectories. We use spatial cross-attention modulation to guide the spatial-temporal placement of objects, all without additional optimization.
Note that our approach, by independently controlling camera movement and object motion, effectively decouples the two, thereby enabling both individual and joint control.}
\label{fig.pipeline}
\end{figure*}
\subsection{Overview} \label{sec.meth.overview}

\paragraph{Task formulation} \label{sec.meth.overview.task}

In this paper, we focus on text-to-video generation with user-directed camera movement and/or object motion. First of all, user should provide a text prompt which may optionally contain one or more object words $O_1,O_2,...O_N$. 
To determine the camera movement, user can specify an x-pan ratio $c_x$, a y-pan ratio $c_y$, and a zoom ratio $c_z$.
To determine the motion of $n$-th object $O_n$ , user needs to specify a starting box $\mathbf{B}^1_n$, an ending box $\mathbf{B}^L_n$ ($L$ is the video length), and an intermediate track $\zeta_n$ connecting $\mathbf{B}^1_n$ and $\mathbf{B}^L_n$, our system then generates a sequence of boxes $[\mathbf{B}^1_n,...,\mathbf{B}^L_n]$ centered along the track $\zeta_n$ via interpolation to define the spatial-temporal journey of the object.
Consequently, our model synthesizes a video that adheres to the prescribed camera movement and/or object motion, creating customized and dynamic visual narrative.


\paragraph{Overall pipeline} \label{sec.meth.overview.pipeline}
Our overall pipeline is illustrated in Figure~\ref{fig.pipeline}. The camera movement is learned in the training stage and the object motion is implemented in the inference stage. 
During the training, we use video samples captured by a stationary camera, which are then augmented to simulate camera movement according to $[c_x, c_y, c_z]$. The augmented videos are subsequently used as input to the U-Net. Additionally, the camera parameters are also encoded and injected into a newly introduced trainable temporal cross-attention layer to condition the camera movement (detailed in Section~\ref{sec.meth.cam}).
During the inference, with trained camera embedder and module, users can specify the camera parameters to control its movement. Concurrently, we incorporate the object motion control in a training-free manner: given the object words from the user's prompt and the corresponding boxes, we modulate the frame-wise and object-wise spatial cross-attention maps to redirect the object spatial-temporal size and location (detailed in Section~\ref{sec.meth.obj}). It is noteworthy that the modulation in inference stage does not involve additional optimization, thus the incremental time and memory cost is negligible.

\subsection{Camera Movement Control} \label{sec.meth.cam}
We choose three types of camera movement: horizontal pan, vertical pan, and zoom, parameterized as a triplet $ \mathbf{c}_\mathrm{cam} =[c_x, c_y, c_z]$ to serve as the camera control signal. 
This allows for quantitative control, a feature not available in previous work \cite{animatediff}, and is simple to use and sufficiently expressive for our needs.

\paragraph{Data construction and augmentation}
\label{sec.meth.cam.training}
Extracting camera movement information from existing video can be computationally expensive since the object motion needs to be identified and filtered out. As such, we propose a self-supervised training approach using camera augmentation driven by $\mathbf{c}_\mathrm{cam}$, thereby bypassing the need for intensive movement annotation.

We first formally define the camera movement parameters. $c_x$ represents the x-pan ratio, and is defined as the total x-shift of the frame center from the first to the last frame relative to the frame width, $c_x>0$ for panning rightward (\eg, $c_x=0.5$ for a half-width right shift). Similarly, $c_y$ is the y-pan ratio, representing the total y-shift of the frame center over the frame height, $c_y>0$ for panning downward. $c_z$ denotes the zoom ratio, defined as the scaling ratio of the last frame relative to the first frame, $c_z>1$ for zooming-in. We set the range of $c_x$,$c_y$ to $[-1,1]$ and $c_z$ to $[0.5,2]$, which are generally sufficient for covering regular camera movement range.

In practice, for given $\mathbf{c}_\mathrm{cam}$, we simulate camera movement by applying shifting and scaling to the cropping window on videos captured with a stationary camera. This data augmentation exploits readily available datasets like MovieShot~\cite{movieshot}. Further details of this process, including pseudo code and sampling scheme of $\mathbf{c}_\mathrm{cam}$ are provided in the supplemental material.

\paragraph{Camera embedding}
\label{sec.meth.cam.emb}
To encode $\mathbf{c}_\mathrm{cam}$ into a camera embedding, we use a camera embedder that includes a Fourier embedder, which is widely used for encoding coordinate-like data \cite{nerf}, and two MLPs. One MLP jointly encodes the panning movement $c_x$, $c_y$, while the other encodes the zooming movement $c_z$.
We empirically found that separately encoding panning and zooming helps the model distinguish between these two distinct types of camera movements effectively, and we validate this design in Section \ref{sec.exp.abl.cam}. The embedding process can be formulated as $\mathbf{e}_{xy}=\mathrm{MLP}_{xy} (\mathcal{F}([c_x,c_y ]))$ ,$\mathbf{e}_z=\mathrm{MLP}_z (\mathcal{F}(c_z ))$, where $\mathcal{F}$ denotes Fourier embedder. Both $\mathbf{e}_{xy}$ and $\mathbf{e}_z$ have the same feature dimensions, By concatenating them, we obtain the camera embedding $\mathbf{e}_\mathrm{cam}=[\mathbf{e}_{xy},\mathbf{e}_z]$, which has a sequence length of two.

\paragraph{Camera module}
\label{sec.meth.cam.module}
We now consider where to inject the camera embedding. Previous studies have highlighted the role of temporal layers in managing temporal transitions \cite{animatediff, motiondirector}. As such, we inject camera control signals via temporal layers. Inspired by the way spatial cross-attention interprets textual information, we introduce new trainable temporal cross-attention layers specifically for interpreting camera information, dubbed as camera modules, which are appended after the existing temporal self-attention layers within each U-Net block of the T2V model, as depicted by the orange box in Figure~\ref{fig.pipeline}. Similar to textual cross-attention, in this module, the queries are mapped from visual frame features $\mathbf{F}$, we separately map the keys and values from panning embedding $\mathbf{e}_{xy}$ and zooming embedding $\mathbf{e}_{z}$ for the same reason stated in the previous section. Through temporal cross-attention, the camera movement is infused into the visual features, which is then added back as a gated residual. We formulate this process as follows:
\begin{equation}
\label{eq.temp_ca1}
\mathbf{F} = \mathbf{F} + \tanh(\alpha) \cdot \text{TempCrossAttn}(\mathbf{F}, \mathbf{e}_\mathrm{cam})
\end{equation}
\begin{equation}    \label{eq.temp_ca2}
\text{TempCrossAttn}(\mathbf{F}, \mathbf{e}_\mathrm{cam}) = \mathrm{Softmax}\left(\frac{{\mathbf{Q}[\mathbf{K}_{xy}, \mathbf{K}_z]^T}}{{\sqrt{d}}}\right) [\mathbf{V}_{xy}, \mathbf{V}_z],
\end{equation}
where $[,]$ denotes concatenation in sequence dimension, $\mathbf{K}_{xy}$,$\mathbf{K}_z$ are key vectors, $\mathbf{V}_{xy}$,$\mathbf{V}_z$ are value vectors mapped from the $\mathbf{e}_{xy}$, $\mathbf{e}_z$ respectively, $d$ is the feature dimension of $\mathbf{Q}$, and $\alpha$ is a learnable scalar initialized as 0, ensuring that the camera movement is gradually learned from the pretrained state.

To learn camera movement while preserving the model's prior knowledge, we freeze the original weights and train only the newly added camera embedder and camera module. These are conditioned on camera movement $\mathbf{c}_{\mathrm{cam}}$, and video caption $c_{\mathrm{txt}}$.
The training employs the diffusion noise-prediction loss:
\begin{equation}   \label{eq.loss}
\mathcal{L}=\mathbb{E}_{\mathbf{x}_0,  \mathbf{c}_\mathrm{cam}, c_\mathrm{txt}, t, \boldsymbol{\epsilon} \sim \mathcal{N}(0, I)}\left[\left\|\boldsymbol{\epsilon}-\boldsymbol{\epsilon}_\theta\left(\mathbf{x}_t, \mathbf{c}_\mathrm{cam}, c_\mathrm{txt}, t \right)\right\|_2^2\right],
\end{equation}
where $\mathbf{x}_0$ is the augmented input sample, $t$ denotes the diffusion timestep, $\mathbf{x}_t=\alpha_t \mathbf{x}_0+ \sigma_t \boldsymbol{\epsilon}$ is the noised sample at $t$, $\alpha_t$ and $\sigma_t$ are time-dependent DDPM hyper-parameters \cite{ddpm}, $\boldsymbol{\epsilon}_\theta$ is the diffusion model parameterized by $\theta$.

\subsection{Object Motion Control} \label{sec.meth.obj}

We choose the bounding box as the control signal for object motion as it aligns best with our method, \ie, modulating attention values within regions defined by boxes. Additionally, boxes are more efficient than dense conditions (\eg, sketch maps require drawing skills) and are more expressive than sparse conditions (\eg, key points lack the specification for object's size).

While it is theoretically possible to train a box-conditioned T2V model similar to GLIGEN~\cite{gligen}. 
However, unlike images, well-annotated video grounding datasets are less accessible, curating and training on large-scale dataset can be labor-intensive and computationally expensive.
To bypass this issue, we opt to fully leverage the inherent priors of pretrained T2V models by steering the diffusion process to our desired result. Previous T2I works have demonstrated the ability to control an object's spatial position by editing cross-attention maps \A{\cite{p2p, ediffi, collagediff, densediff, directeddiff,layoutgui}}. Similarly, we employ the spatial cross-attention modulation in T2V model for object motion crafting.

In cross-attention layers, the query features $\mathbf{Q}$ are derived from visual tokens, the key $\mathbf{K}$ and value features $\mathbf{V}$ are mapped from textual tokens. $\mathbf{QK}^\top$ constitutes an attention map, where the value at index $[i, j]$ reflects the response of the i-th image token feature to the j-th textual token feature. We modulate the attention map $\mathbf{QK}^\top$ as follows:
\begin{equation}   \label{eq.attn_qkv}
\text{CrossAttnModulate}(\mathbf{Q, K, V}) =  \text{Softmax}\left(\frac{\mathbf{QK}^\top + \lambda \mathbf{S}}{\sqrt{d}}\right) \mathbf{V},
\end{equation}
where $\lambda$ represents modulation strength,  $d$ is the feature dimension of $\mathbf{Q}$, and $\mathbf{S}$ is the modulation term of the same size as $\mathbf{QK}^\top$. It comprises two types of modulation: amplification and suppression.

\paragraph{Attention amplification} Considering the $n$-th object in the $k$-th frame, enclosed by the bounding box $\mathbf{B}_n^k$, since we aim to increase the probability of the object's presence in this region, we could amplify the attention values for the corresponding object words (indexed as $\mathbf{T}_n$ in the prompt) within the area $\mathbf{B}_n^k$. Note that if there exists a background word, we treat it in the same way, and its region is the complement of the union of all the objects' regions. Following the conclusion from DenseDiff \cite{densediff}, the scale of this amplification should be inversely related to the area of $\mathbf{B}_n^k$, \ie, smaller box area are subject to a larger increase in attention. Since our attention amplification is performed on box-shaped regions, which does not align with the object's natural contours, we confine the amplification to the early stages (for timesteps $t \geq \tau$, $\tau$ is the cut-off timestep), as the early stage mainly focuses on generating coarse layouts. For $t < \tau$, we relax this control to enable the diffusion process to gradually refine the shape and appearance details.

\paragraph{Attention suppression} To mitigate the influence of irrelevant words on the specified region and prevent the unintended dispersion of object features to other areas, we suppress attention values for unmatched query-key token pairs (except start token <sos> and end token <eos> otherwise the video quality would be compromised). Different from attention amplification, attention suppression is applied throughout the entire sampling process to prevent mutual semantic interference, an potential issue in multi-object generation scenarios where the semantics of one object might inadvertently bleed into another. We will present the results and analysis in the ablation studies (Section~\ref{sec.exp.abl}).

Formally, the attention modulation term for the $n$-th object in the $k$-th frame $\mathbf{S}_n^k [i,j]$ is formulated as:
\begin{equation}   \label{eq.attn_s}
\mathbf{S}_n^k [i,j]=
\begin{cases} 
1-\frac{|\mathbf{B}_n^k|}{|\mathbf{QK}^\top|}, & \text{if } i \in \mathbf{B}_n^k \text{ and } j \in \mathbf{T}_n \text{ and } t \geq \tau \\
0, & \text{if } i \in \mathbf{B}_n^k \text{ and } j \in \mathbf{T}_n \text{ and } t < \tau \\
-\infty, & \text{otherwise}
\end{cases} 
\end{equation}
where $|\mathbf{X}|$ denotes the number of elements in matrix $\mathbf{X}$. We perform such modulation for each object in every frame so that the complete spatial-temporal object trajectory can be determined. Note that although this modulation is independently performed in each frame, we observe that the generated videos remain continuous, thanks to the pretrained temporal layers which maintains temporal continuity.

\section{Experiment} \label{sec.exp}
\subsection{Experimental Setup} \label{sec.exp.setup}

\paragraph{Implementation details}  \label{sec.exp.setup.impl}
We adopt pretrained Zeroscope T2V model \cite{modelscope} as our base model, integrating our proposed trainable camera embedder and module to facilitate camera movement learning, please refer to supplementary materials for training details.
During the inference, we use DDIM sampler \cite{ddim} with $T=50$ sampling steps and a classifier-free guidance scale of 9 \cite{cfg}. The default attention control weight $\lambda$ and cut-off timestep $\tau$ are 25 and $0.95T$ respectively. The output video size is 320×512×24.

\paragraph{Datasets}  \label{sec.exp.setup.datasets}
For camera movement training, we use a subset from MovieShot \cite{movieshot}, which contains 22k static-shot movie trailers, \ie, the camera is fixed but the subject is flexible to move, ensuring that the training samples are devoid of original camera movement. Despite the limited number and category of the training data, our trained camera module is still able to adapt to general scenes.
For camera control evaluation, we collected 200 scene prompts from the prompt set provided by~\cite{Chivileva2023}. For object control evaluation, we curated a benchmark of 200 box-prompt pairs, comprising varied box sizes, locations, and trajectories, with prompts primarily focusing on natural animals and objects.

\paragraph{Metrics}
\label{sec.exp.setup.metrics}

\noindent (1) To assess video generation quality, we employ FID-vid \cite{fid} and FVD \cite{fvd}. The reference set consist of 2048 videos from MSRVTT \cite{msrvtt} for the camera control task and 800 videos from AnimalKingdom \cite{animalkingdom} for the object control task. 
\noindent (2) \R{To evaluate camera movement control, we introduce the flow error metric}. We utilize VideoFlow \cite{videoflow}, a state-of-the-art optical flow model, to extract flow maps from the generated videos. These are then compared against the ground truth flow maps, which are derived from the given camera movement parameters.
\noindent (3) To measure the object-prompt alignment in object control task, we uniformly extract 8 frames per video sample and calculate the CLIP image-text similarity (CLIP-sim) \cite{clipscore} within the box area, with a templated prompt ``a photo of <obj>'', where <obj> corresponds to the object phrase. 
\noindent \R{(4) To measure the object-box alignment, we employ Grounding DINO \cite{groundingdino} to detect object boxes in generated videos. We then calculate the mean Intersection over Union (mIoU) against the input boxes and compute the average precision score at the 0.5 IoU threshold (AP50).}

\paragraph{Baselines}  \label{sec.exp.setup.baseline}
We compare our method with recent diffusion-based T2V models with the camera movement or object motion controllability, including AnimateDiff \cite{animatediff} (for camera movement), Peekaboo \cite{peekaboo} (for object motion), and VideoComposer \cite{videocomposer} (for both).

\subsection{Camera Movement Control} \label{sec.exp.cam}

For camera movement control, we conduct comparisons with AnimateDiff and VideoComposer. For AnimateDiff, we use official pretrained LoRA motion modules, each dedicated to a specific type of camera movement but lacking support for precise control. For VideoComposer, we hand-craft a motion vector map based on the camera movement parameters, as demonstrated in its paper.

\paragraph{Qualitative comparison}
\label{sec.exp.cam.quali}

\begin{figure*}
\centering
   \includegraphics[width=\linewidth]{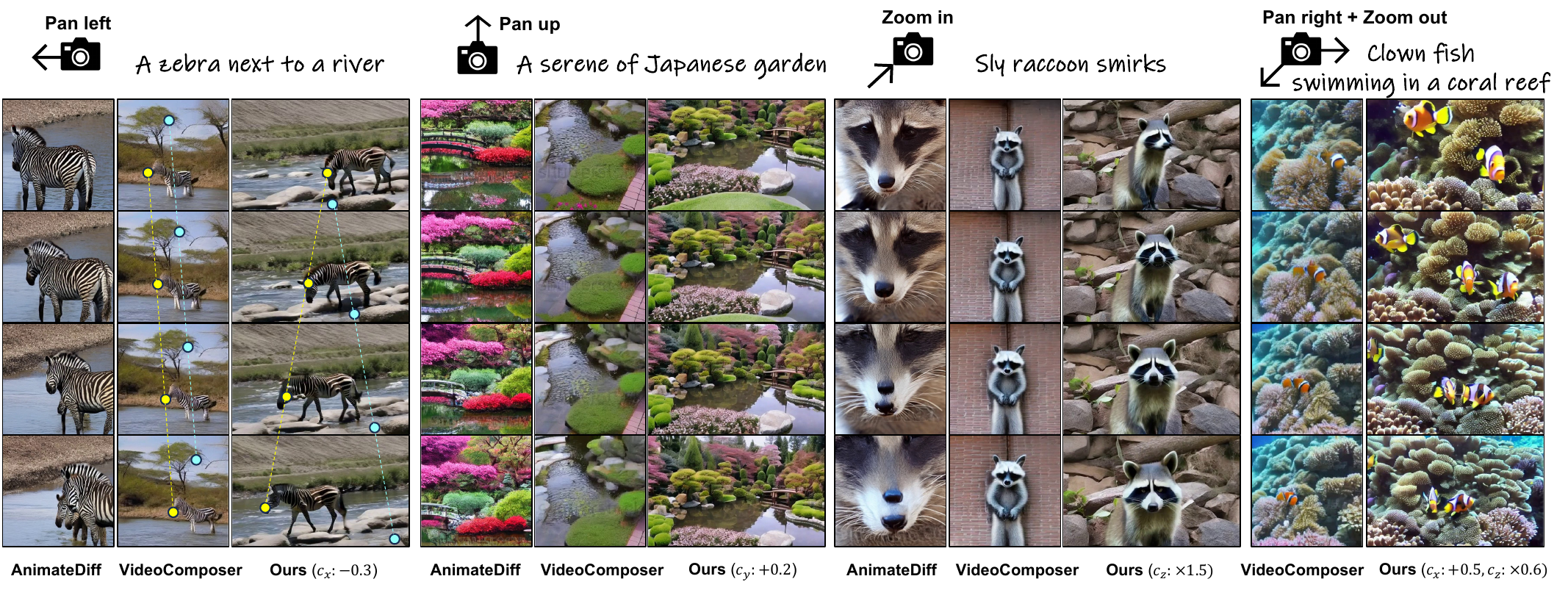}
   \caption{Qualitative comparison on camera movement control with related baselines. Our results in the third column show that the object motion (\textcolor{yellow}{yellow} lines) can be independent from the camera movement (\textcolor{cyan}{cyan} lines) unlike results by VideoComposer~\cite{videocomposer} in the second column.
   }
\label{fig.cam}
\end{figure*}

We present side-by-side visual comparison with baselines in Figure~\ref{fig.cam}. As can be seen, all the methods are capable of generating videos with the single type of camera movement, but AnimateDiff does not support hybrid camera movement (\eg, pan+zoom) since its loaded motion module is dedicated to one type of camera movement only, while our method and VideoComposer can combine or switch the camera movement by altering the motion input, without the need for re-loading extra modules. 
In terms of precise control, both our method and VideoComposer can quantitatively control the camera speed. 
\R{Specifically, VideoComposer~\cite{videocomposer} requires a sequence of motion maps as input, while ours only requires three camera parameters.}
Moreover, in terms of disentanglement, our method's camera control does not impact foreground objects, as we do not impose any motion constraints on them. In contrast, VideoComposer employs a global motion vector map, which often binds objects together with background movement. As shown in the 3rd column of Figure~\ref{fig.cam}, the zebra in our results exhibits its independent motion from the camera, whereas in VideoComposer's results (the 2nd column), the zebra is tied to the camera movement, so does the fish in the last 2nd column. Finally, our results also exhibit higher visual quality, a testament to the superiority of our base model.

\paragraph{Quantitative comparison}
\label{sec.exp.cam.quanti}

We report FVD, FID-vid, and Flow error in Table~\ref{tab.cam.quanti}. Note that AnimateDiff is excluded from the flow error comparison due to its lack of quantitative control. Our results achieve the best FVD and FID-vid scores, indicating superior visual quality compared to baselines, and show more precise camera control, evidenced by a lower flow error.

\begin{table}
\centering
\caption{Quantitative comparison for camera movement control evaluation.}
\label{tab.cam.quanti}
\setlength{\tabcolsep}{1.5mm}
\begin{tabular}{@{}l|cccc@{}}
\toprule
              & FVD $\downarrow$ & FID-vid$\downarrow$ & Flow error$\downarrow$   
              \\ \midrule
AnimateDiff   & 1685.40         & 82.57           & -             \\
VideoComposer & 1230.57         & 82.14           & 0.74          \\
\sysname (ours)          & \textbf{888.91} & \textbf{48.96}  & \textbf{0.46} \\ 
\bottomrule
\end{tabular}
\end{table}

\subsection{Object Motion Control}
\label{sec.exp.obj}

\begin{figure*}
\begin{center}
   \includegraphics[width=\linewidth]{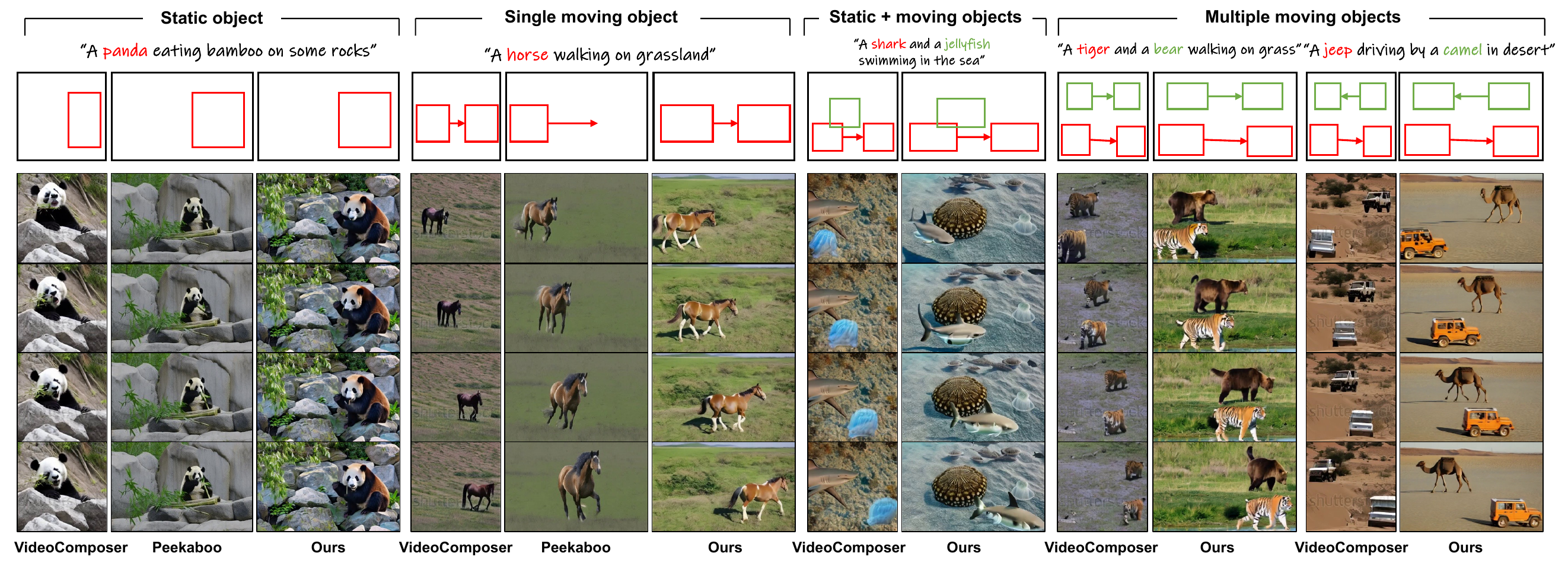}
\end{center}
   \caption{Qualitative comparison on object motion control with related baselines. Our method excels in handling cases involving more than one object.}
\label{fig.obj}
\end{figure*}

For object motion control, we compare with VideoComposer~\cite{videocomposer} \R{and Peekaboo~\cite{peekaboo}}. To enable VideoComposer generating object motion via boxes, we first convert object box sequences into dense flow maps, which are then processed into motion vector maps compatible with its input format. Peekaboo's visual results are taken from their official website.

\paragraph{Qualitative comparison}
\label{sec.exp.obj.quali}

We present visual comparison with related baselines in Figure~\ref{fig.obj}. For static object generation, VideoComposer fails to generate the object in desired location (see the panda in the first column), without any motion hint, it works like a vanilla T2V model. While all methods are capable of generating a single moving object, challenges arise in multiple moving objects scenarios. 
Peekaboo is excluded from this comparison as its code is not implemented for multiple objects. 
VideoComposer does not support specifying individual motion for each object unlike our method (see the shark and jellyfish examples in the 7th and 8th columns). Moreover, its lack of explicit binding between objects and motion leads to two extra issues: semantic mixing and absence. Semantic mixing refers to the blending of one object's semantics with another. This is exemplified in the 9th column, where tiger's texture leaks into bear. Semantic absence occurs when an object does not appear as anticipated, a known issue in T2I/T2V models \cite{attnexcite}. For instance, in the 11th column, the expected camel is missing, replaced instead by a jeep. In contrast, our method effectively addresses these issues through ad-hoc attention modulation for each object, facilitating easier control over multiple objects' motion.

\paragraph{Quantitative comparison}
\label{sec.exp.obj.quanti}
We report quality metrics (FVD, FID-vid) and grounding metrics (CLIP-sim, \R{mIoU, AP50}) in Table~\ref{tab.obj.quanti}. 
\R{In terms of quality, our method is comparable to Peekaboo, as both utilize the same superior model that outperforms VideoComposer's. For object control, our method slightly surpasses VideoComposer and significantly exceeds Peekaboo by additionally incorporating attention amplification, in contrast to Peekaboo's reliance on attention masking alone. We believe the use of amplification plays important role in improving grounding ability, as demonstrated in our ablation study (Section~\ref{sec.exp.abl}).}

\begin{table}
\small
\centering
\caption{Quantitative comparison for object motion control evaluation.}
\label{tab.obj.quanti}
\setlength{\tabcolsep}{0.5mm}
\begin{tabular}{@{}l|ccccc@{}}
\toprule
    & FVD $\downarrow$ & FID-vid$\downarrow$ & CLIP-sim$\uparrow$ & \R{mIoU (\%) $\uparrow$} & \R{AP50 (\%)$\uparrow$}  \\
\midrule
VideoComposer   & 1620.83        & 90.57          & 27.35           &  \R{45.24}    & \R{31.01}            \\
\R{Peekaboo}    & \R{1384.62}   & \R{44.49}       & \R{27.03}         &  \R{36.55}    & \R{18.77}   \\
\sysname & \textbf{1300.86}    & \textbf{43.55}      & \textbf{27.63} &  \R{\textbf{47.83}}    & \R{\textbf{31.33}}    \\
\bottomrule
\end{tabular}
\end{table}

\subsection{Joint Control of Camera Movement and Object Motion}
\label{sec.exp.cam_obj}

\definecolor{yellow}{wave}{590}
\begin{figure*}
\begin{center}
   \includegraphics[width=0.98\linewidth]{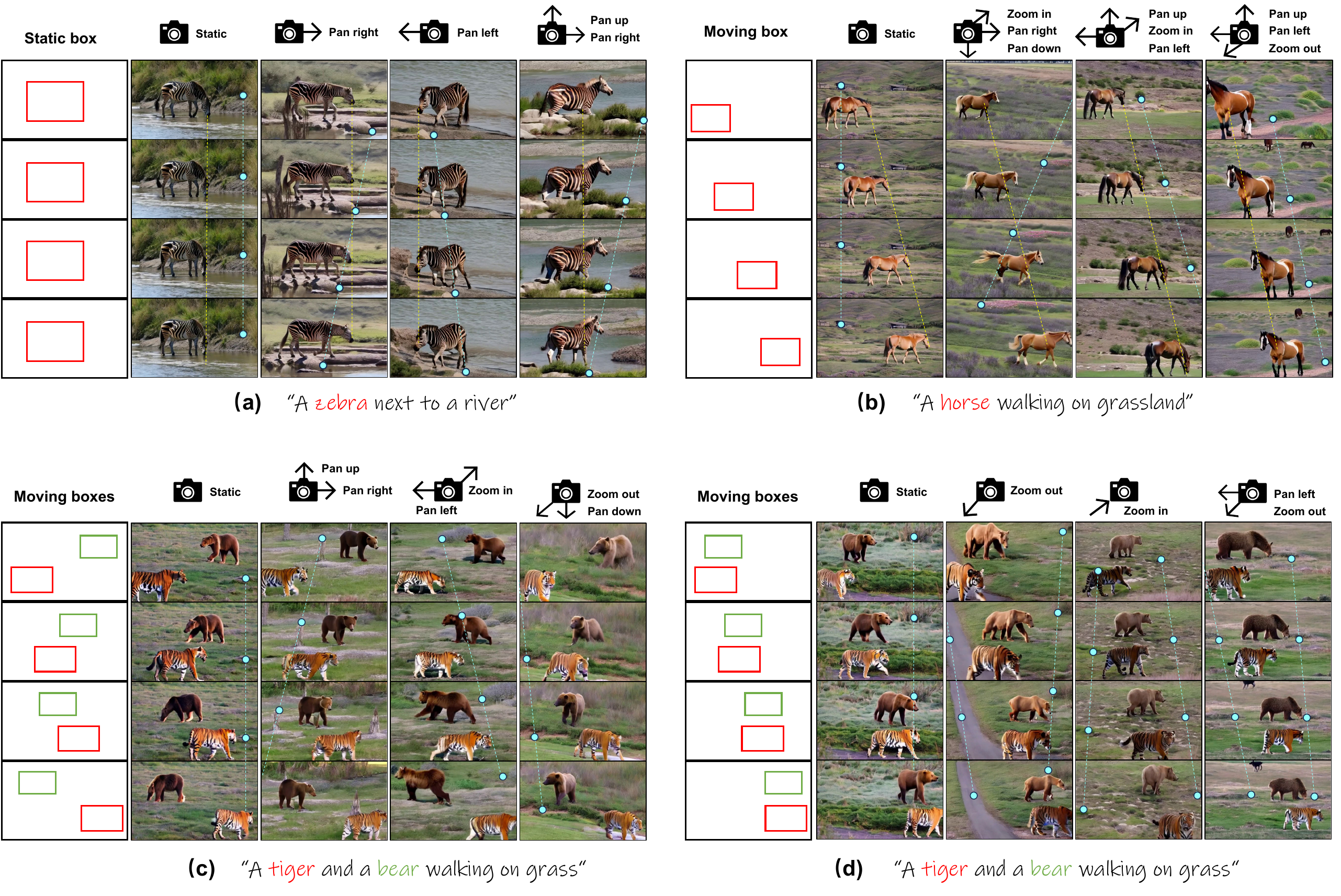}
\end{center}
   \caption{Joint control of object motion and camera movement. Given the same box sequence, by setting different camera movement parameters, our approach is capable of synthesizing videos that exhibit a diverse combination of foreground motion (\textcolor{yellow}{yellow} lines) and background motion (\textcolor{cyan}{cyan} lines). The user can create a well-defined overall video motion by distinctly specifying both object motion and camera movement using our method.}
\label{fig.cam_obj}
\end{figure*}

\sysname features in jointly supporting the control of both camera movement and object motion, we demonstrate such capability in Figure~\ref{fig.cam_obj}. Given the same box sequence, our method can generate videos with varied combination of foreground-background motions. For example, Figure~\ref{fig.cam_obj}(a) illustrates that a static box does not always imply a static object, by setting different camera movements, our system can generate videos of a zebra standing still (2nd column), walking right (3rd column), or walking left (4th column). Similarly, Figure~\ref{fig.cam_obj}(b) suggests that a moving box does not necessarily indicate that the object itself is in motion, it could be stationary in its position while the camera is moving (last column). Existing works focused only on object often fail to differentiate between the object's inherent motion and apparent motion induced by camera movement. In contrast, our method enables users to distinctly specify both camera movement and object motion, offering enhanced flexibility in defining overall motion patterns. More examples are provided in Figure~\ref{fig.extra} and our project page.

\subsection{Ablation Study} \label{sec.exp.abl}
We conduct ablation studies to evaluate several key components of our work.

\begin{figure}
\centering
   \includegraphics[width=\linewidth]{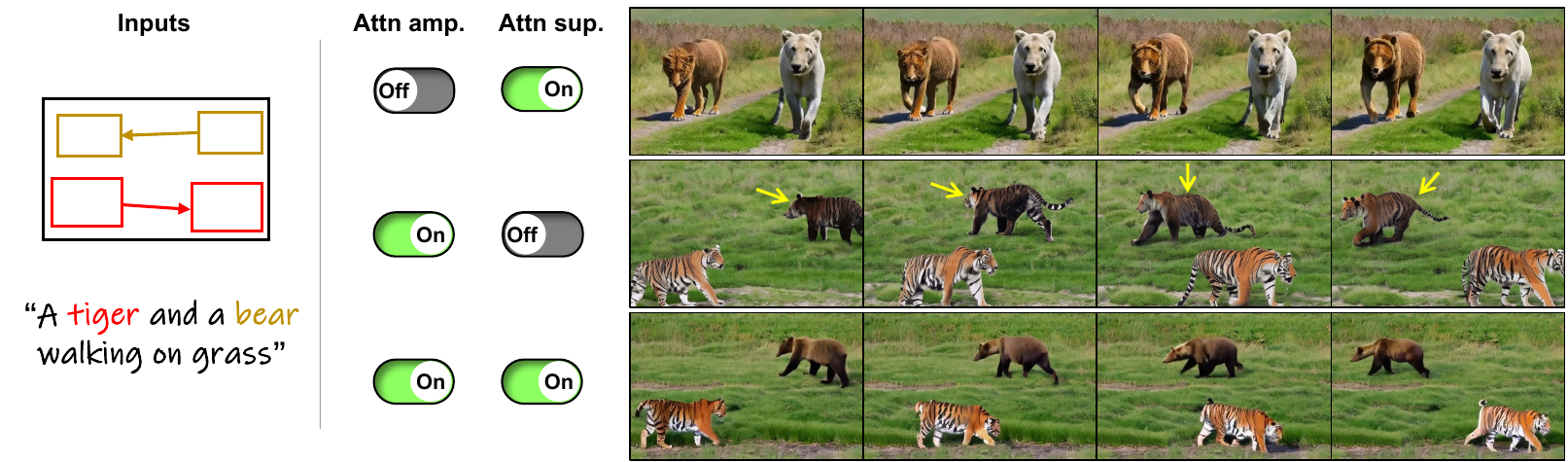}
   \caption{Effect of attention amplification and suppression. Without amplification (the first row), the objects do not adhere to boxes; Without suppression (the second row), tiger’s texture mistakenly leaks into the bear’s body. These issues are resolved with both enabled (the third row).}
\label{fig.abl.attn}
\end{figure}

\paragraph{Attention amplification} This is crucial for object localization, the absence of attention amplification results in a decrease of grounding ability, \ie, the object is less likely to follow the boxes, as shown in the first row in Figure~\ref{fig.abl.attn}, and a decrease of metrics in Table \ref{tab.abl.attn}.

\paragraph{Attention suppression} This is introduced to mitigate the unintended semantic mixing in multi-object scenarios, particularly when objects share similar characteristics. Since our attention amplification is applied only in the initial steps, and this constraint is subsequently relaxed. Without suppression, object A’s prompt feature can also attend to object B’s region, leading to semantic overlap. As shown in second row of Figure~\ref{fig.abl.attn}, where the tiger’s texture erroneously appears on the bear’s body. The third row shows that this issue can be resolved by enabling the attention suppression.

\paragraph{Camera embedding design} \label{sec.exp.abl.cam}
To assess the effectiveness of separately encoding panning ($c_x$, $c_y$) and zooming ($c_z$) movements in camera control as detailed in Section~\ref{sec.meth.cam.emb}, we contrast this with a joint encoding approach. Here, $[c_x, c_y, c_z]$ are encoded into a single camera embedding vector using a shared MLP, followed by shared key-value projection matrix in the camera module.
We train and evaluate the model with the same setting, we observed a reduced ability in camera movement control, with flow error increasing from 0.46 to 1.68. This underscores the advantages of separate encoding for distinct types of camera movements.


\begin{table}
\centering
\caption{Quantitative evaluation of attention amplification and suppression.}
\label{tab.abl.attn}
\begin{tabular}{@{}cc|ccc@{}}
\toprule
Attn amp. & Attn sup. & CLIP-sim $\uparrow$ & \R{mIoU (\%)$\uparrow$} & \R{AP50 (\%)$\uparrow$} \\
\midrule
$\times$ & $\checkmark$     & 25.82               &  \R{15.35}     &         \R{3.46} \\
$\checkmark$ & $\times$     & 27.49              &   \R{38.87}     &         \R{10.25}\\
$\checkmark$ & $\checkmark$ & \textbf{27.63} & \R{\textbf{47.83}}    & \R{\textbf{31.33}} \\
\bottomrule
\end{tabular}
\end{table}

\section{Limitations}  \label{sec.limit}
We consider several limitations of our method.

\begin{figure}[t]
\begin{center}
   \includegraphics[width=\linewidth]{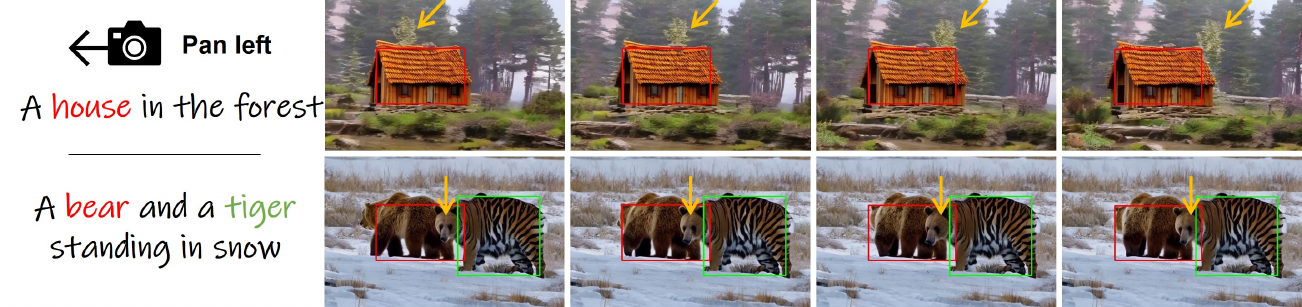}
\end{center}
   \caption{Limitations of our method. Top: conflicting inputs can lead to unreal results - a moving house. Bottom: Overlapping boxes may lead to object interfere - tiger with a bear head.}
\label{fig.limit}
\end{figure}

\noindent (1) For joint control, while our method provides disentangled control over object and camera motion, conflicts can sometimes arise in the inputs. For instance, in the top row of Figure~\ref{fig.limit}, we attempt to maintain a static object (house) within a static box while simultaneously panning the camera to the left. Given these conflicting signals, our method ends up generating a moving house, which is unrealistic. This necessitates careful and reasonable user interaction.

\noindent (2) In camera control, due to the camera augmentation technique used in our method, which currently involves only 2D-panning and zooming, this limits the system's ability to produce complex 3D camera movements that are out of this scope, \eg, our method cannot generate camera movements like "panning around an object". To overcome this constraint, we consider envisaging the adoption of more sophisticated augmentation algorithms, or curating a synthetic video dataset from a rendering engine given the camera movements, which we will leave in our future work.

\noindent (3) In object control, another issue arises when handling colliding boxes. In scenarios like the one depicted in the bottom row of Figure~\ref{fig.limit}, where two boxes overlap, the semantics of one object (the bear) can interfere with another (the tiger). This issue can be mitigated by modulating attention on an adaptively auto-segmented region during the diffusion sampling process, rather than relying on the initial box region.

\section{Conclusion}
\label{sec.conclusion}

In this work, we propose \sysname, a text-to-video framework that addresses the previously unmet need for independent and user-directed control over camera movement and object motion. Our approach effectively decouples these two elements by integrating a self-supervised training scheme for temporal cross-attention layers tailored for camera movement control, with a training-free modulation for spatial cross-attention dedicated to object motion control. Experimental evaluations demonstrate the capability of our approach in separate and joint control of camera movement and object motion. This positions \sysname as an efficient and flexible tool for creative video synthesis with customized motion.


\begin{acks}

This work was supported by a GRF grant (Project No. CityU 11208123) from the Research Grants Council (RGC) of Hong Kong, and research funding from Kuaishou Technology. 
\end{acks}

\bibliographystyle{ACM-Reference-Format}
\bibliography{mybib}


\begin{thebibliography}{74}


\ifx \showCODEN    \undefined \def \showCODEN     #1{\unskip}     \fi
\ifx \showDOI      \undefined \def \showDOI       #1{#1}\fi
\ifx \showISBNx    \undefined \def \showISBNx     #1{\unskip}     \fi
\ifx \showISBNxiii \undefined \def \showISBNxiii  #1{\unskip}     \fi
\ifx \showISSN     \undefined \def \showISSN      #1{\unskip}     \fi
\ifx \showLCCN     \undefined \def \showLCCN      #1{\unskip}     \fi
\ifx \shownote     \undefined \def \shownote      #1{#1}          \fi
\ifx \showarticletitle \undefined \def \showarticletitle #1{#1}   \fi
\ifx \showURL      \undefined \def \showURL       {\relax}        \fi
\providecommand\bibfield[2]{#2}
\providecommand\bibinfo[2]{#2}
\providecommand\natexlab[1]{#1}
\providecommand\showeprint[2][]{arXiv:#2}

\bibitem[Balaji et~al\mbox{.}(2022)]%
        {ediffi}
\bibfield{author}{\bibinfo{person}{Yogesh Balaji}, \bibinfo{person}{Seungjun Nah}, \bibinfo{person}{Xun Huang}, \bibinfo{person}{Arash Vahdat}, \bibinfo{person}{Jiaming Song}, \bibinfo{person}{Karsten Kreis}, \bibinfo{person}{Miika Aittala}, \bibinfo{person}{Timo Aila}, \bibinfo{person}{Samuli Laine}, \bibinfo{person}{Bryan Catanzaro}, {et~al\mbox{.}}} \bibinfo{year}{2022}\natexlab{}.
\newblock \showarticletitle{ediffi: Text-to-image diffusion models with an ensemble of expert denoisers}.
\newblock \bibinfo{journal}{\emph{arXiv preprint arXiv:2211.01324}} (\bibinfo{year}{2022}).
\newblock


\bibitem[Blattmann et~al\mbox{.}(2023a)]%
        {svd}
\bibfield{author}{\bibinfo{person}{Andreas Blattmann}, \bibinfo{person}{Tim Dockhorn}, \bibinfo{person}{Sumith Kulal}, \bibinfo{person}{Daniel Mendelevitch}, \bibinfo{person}{Maciej Kilian}, \bibinfo{person}{Dominik Lorenz}, \bibinfo{person}{Yam Levi}, \bibinfo{person}{Zion English}, \bibinfo{person}{Vikram Voleti}, \bibinfo{person}{Adam Letts}, {et~al\mbox{.}}} \bibinfo{year}{2023}\natexlab{a}.
\newblock \showarticletitle{Stable video diffusion: Scaling latent video diffusion models to large datasets}.
\newblock \bibinfo{journal}{\emph{arXiv preprint arXiv:2311.15127}} (\bibinfo{year}{2023}).
\newblock


\bibitem[Blattmann et~al\mbox{.}(2023b)]%
        {videoldm}
\bibfield{author}{\bibinfo{person}{Andreas Blattmann}, \bibinfo{person}{Robin Rombach}, \bibinfo{person}{Huan Ling}, \bibinfo{person}{Tim Dockhorn}, \bibinfo{person}{Seung~Wook Kim}, \bibinfo{person}{Sanja Fidler}, {and} \bibinfo{person}{Karsten Kreis}.} \bibinfo{year}{2023}\natexlab{b}.
\newblock \showarticletitle{Align your latents: High-resolution video synthesis with latent diffusion models}. In \bibinfo{booktitle}{\emph{Proceedings of the IEEE/CVF Conference on Computer Vision and Pattern Recognition}}. \bibinfo{pages}{22563--22575}.
\newblock


\bibitem[Ceylan et~al\mbox{.}(2023)]%
        {pix2video}
\bibfield{author}{\bibinfo{person}{Duygu Ceylan}, \bibinfo{person}{Chun-Hao~P Huang}, {and} \bibinfo{person}{Niloy~J Mitra}.} \bibinfo{year}{2023}\natexlab{}.
\newblock \showarticletitle{Pix2video: Video editing using image diffusion}. In \bibinfo{booktitle}{\emph{Proceedings of the IEEE/CVF International Conference on Computer Vision}}. \bibinfo{pages}{23206--23217}.
\newblock


\bibitem[Chai et~al\mbox{.}(2023)]%
        {stablevideo}
\bibfield{author}{\bibinfo{person}{Wenhao Chai}, \bibinfo{person}{Xun Guo}, \bibinfo{person}{Gaoang Wang}, {and} \bibinfo{person}{Yan Lu}.} \bibinfo{year}{2023}\natexlab{}.
\newblock \showarticletitle{Stablevideo: Text-driven consistency-aware diffusion video editing}. In \bibinfo{booktitle}{\emph{Proceedings of the IEEE/CVF International Conference on Computer Vision}}. \bibinfo{pages}{23040--23050}.
\newblock


\bibitem[Chang et~al\mbox{.}(2023)]%
        {magicdance}
\bibfield{author}{\bibinfo{person}{Di Chang}, \bibinfo{person}{Yichun Shi}, \bibinfo{person}{Quankai Gao}, \bibinfo{person}{Jessica Fu}, \bibinfo{person}{Hongyi Xu}, \bibinfo{person}{Guoxian Song}, \bibinfo{person}{Qing Yan}, \bibinfo{person}{Xiao Yang}, {and} \bibinfo{person}{Mohammad Soleymani}.} \bibinfo{year}{2023}\natexlab{}.
\newblock \showarticletitle{MagicDance: Realistic Human Dance Video Generation with Motions \& Facial Expressions Transfer}.
\newblock \bibinfo{journal}{\emph{arXiv preprint arXiv:2311.12052}} (\bibinfo{year}{2023}).
\newblock


\bibitem[Chefer et~al\mbox{.}(2023)]%
        {attnexcite}
\bibfield{author}{\bibinfo{person}{Hila Chefer}, \bibinfo{person}{Yuval Alaluf}, \bibinfo{person}{Yael Vinker}, \bibinfo{person}{Lior Wolf}, {and} \bibinfo{person}{Daniel Cohen-Or}.} \bibinfo{year}{2023}\natexlab{}.
\newblock \showarticletitle{Attend-and-excite: Attention-based semantic guidance for text-to-image diffusion models}.
\newblock \bibinfo{journal}{\emph{ACM Transactions on Graphics (TOG)}} \bibinfo{volume}{42}, \bibinfo{number}{4} (\bibinfo{year}{2023}), \bibinfo{pages}{1--10}.
\newblock


\bibitem[Chen et~al\mbox{.}(2024)]%
        {layoutgui}
\bibfield{author}{\bibinfo{person}{Minghao Chen}, \bibinfo{person}{Iro Laina}, {and} \bibinfo{person}{Andrea Vedaldi}.} \bibinfo{year}{2024}\natexlab{}.
\newblock \showarticletitle{Training-free layout control with cross-attention guidance}. In \bibinfo{booktitle}{\emph{Proceedings of the IEEE/CVF Winter Conference on Applications of Computer Vision}}. \bibinfo{pages}{5343--5353}.
\newblock


\bibitem[Chen et~al\mbox{.}(2023a)]%
        {mcdiff}
\bibfield{author}{\bibinfo{person}{Tsai-Shien Chen}, \bibinfo{person}{Chieh~Hubert Lin}, \bibinfo{person}{Hung-Yu Tseng}, \bibinfo{person}{Tsung-Yi Lin}, {and} \bibinfo{person}{Ming-Hsuan Yang}.} \bibinfo{year}{2023}\natexlab{a}.
\newblock \showarticletitle{Motion-Conditioned Diffusion Model for Controllable Video Synthesis}.
\newblock \bibinfo{journal}{\emph{arXiv preprint arXiv:2304.14404}} (\bibinfo{year}{2023}).
\newblock


\bibitem[Chen et~al\mbox{.}(2023b)]%
        {controlavideo}
\bibfield{author}{\bibinfo{person}{Weifeng Chen}, \bibinfo{person}{Jie Wu}, \bibinfo{person}{Pan Xie}, \bibinfo{person}{Hefeng Wu}, \bibinfo{person}{Jiashi Li}, \bibinfo{person}{Xin Xia}, \bibinfo{person}{Xuefeng Xiao}, {and} \bibinfo{person}{Liang Lin}.} \bibinfo{year}{2023}\natexlab{b}.
\newblock \showarticletitle{Control-A-Video: Controllable Text-to-Video Generation with Diffusion Models}.
\newblock \bibinfo{journal}{\emph{arXiv preprint arXiv:2305.13840}} (\bibinfo{year}{2023}).
\newblock


\bibitem[Chivileva et~al\mbox{.}(2023)]%
        {Chivileva2023}
\bibfield{author}{\bibinfo{person}{Iya Chivileva}, \bibinfo{person}{Philip Lynch}, \bibinfo{person}{Tomas Ward}, {and} \bibinfo{person}{Alan Smeaton}.} \bibinfo{year}{2023}\natexlab{}.
\newblock \showarticletitle{{Text prompts and videos generated using 5 popular Text-to-Video models plus quality metrics including user quality assessments}}.
\newblock  (\bibinfo{date}{10} \bibinfo{year}{2023}).
\newblock
\urldef\tempurl%
\url{https://doi.org/10.6084/m9.figshare.24078045.v2}
\showDOI{\tempurl}


\bibitem[Deng et~al\mbox{.}(2023)]%
        {dragvideo}
\bibfield{author}{\bibinfo{person}{Yufan Deng}, \bibinfo{person}{Ruida Wang}, \bibinfo{person}{Yuhao Zhang}, \bibinfo{person}{Yu-Wing Tai}, {and} \bibinfo{person}{Chi-Keung Tang}.} \bibinfo{year}{2023}\natexlab{}.
\newblock \showarticletitle{DragVideo: Interactive Drag-style Video Editing}.
\newblock \bibinfo{journal}{\emph{arXiv preprint arXiv:2312.02216}} (\bibinfo{year}{2023}).
\newblock


\bibitem[Esser et~al\mbox{.}(2023)]%
        {gen1}
\bibfield{author}{\bibinfo{person}{Patrick Esser}, \bibinfo{person}{Johnathan Chiu}, \bibinfo{person}{Parmida Atighehchian}, \bibinfo{person}{Jonathan Granskog}, {and} \bibinfo{person}{Anastasis Germanidis}.} \bibinfo{year}{2023}\natexlab{}.
\newblock \showarticletitle{Structure and content-guided video synthesis with diffusion models}. In \bibinfo{booktitle}{\emph{Proceedings of the IEEE/CVF International Conference on Computer Vision}}. \bibinfo{pages}{7346--7356}.
\newblock


\bibitem[Feng et~al\mbox{.}(2023)]%
        {dreamoving}
\bibfield{author}{\bibinfo{person}{Mengyang Feng}, \bibinfo{person}{Jinlin Liu}, \bibinfo{person}{Kai Yu}, \bibinfo{person}{Yuan Yao}, \bibinfo{person}{Zheng Hui}, \bibinfo{person}{Xiefan Guo}, \bibinfo{person}{Xianhui Lin}, \bibinfo{person}{Haolan Xue}, \bibinfo{person}{Chen Shi}, \bibinfo{person}{Xiaowen Li}, {et~al\mbox{.}}} \bibinfo{year}{2023}\natexlab{}.
\newblock \showarticletitle{DreaMoving: A Human Video Generation Framework based on Diffusion Models}.
\newblock \bibinfo{journal}{\emph{arXiv e-prints}} (\bibinfo{year}{2023}), \bibinfo{pages}{arXiv--2312}.
\newblock


\bibitem[Gal et~al\mbox{.}(2022)]%
        {txtinv}
\bibfield{author}{\bibinfo{person}{Rinon Gal}, \bibinfo{person}{Yuval Alaluf}, \bibinfo{person}{Yuval Atzmon}, \bibinfo{person}{Or Patashnik}, \bibinfo{person}{Amit~H Bermano}, \bibinfo{person}{Gal Chechik}, {and} \bibinfo{person}{Daniel Cohen-Or}.} \bibinfo{year}{2022}\natexlab{}.
\newblock \showarticletitle{An image is worth one word: Personalizing text-to-image generation using textual inversion}.
\newblock \bibinfo{journal}{\emph{arXiv preprint arXiv:2208.01618}} (\bibinfo{year}{2022}).
\newblock


\bibitem[Geyer et~al\mbox{.}(2023)]%
        {tokenflow}
\bibfield{author}{\bibinfo{person}{Michal Geyer}, \bibinfo{person}{Omer Bar-Tal}, \bibinfo{person}{Shai Bagon}, {and} \bibinfo{person}{Tali Dekel}.} \bibinfo{year}{2023}\natexlab{}.
\newblock \showarticletitle{Tokenflow: Consistent diffusion features for consistent video editing}.
\newblock \bibinfo{journal}{\emph{arXiv preprint arXiv:2307.10373}} (\bibinfo{year}{2023}).
\newblock


\bibitem[Gu et~al\mbox{.}(2023)]%
        {videoswap}
\bibfield{author}{\bibinfo{person}{Yuchao Gu}, \bibinfo{person}{Yipin Zhou}, \bibinfo{person}{Bichen Wu}, \bibinfo{person}{Licheng Yu}, \bibinfo{person}{Jia-Wei Liu}, \bibinfo{person}{Rui Zhao}, \bibinfo{person}{Jay~Zhangjie Wu}, \bibinfo{person}{David~Junhao Zhang}, \bibinfo{person}{Mike~Zheng Shou}, {and} \bibinfo{person}{Kevin Tang}.} \bibinfo{year}{2023}\natexlab{}.
\newblock \showarticletitle{VideoSwap: Customized Video Subject Swapping with Interactive Semantic Point Correspondence}.
\newblock \bibinfo{journal}{\emph{arXiv preprint arXiv:2312.02087}} (\bibinfo{year}{2023}).
\newblock


\bibitem[Guo et~al\mbox{.}(2023)]%
        {animatediff}
\bibfield{author}{\bibinfo{person}{Yuwei Guo}, \bibinfo{person}{Ceyuan Yang}, \bibinfo{person}{Anyi Rao}, \bibinfo{person}{Yaohui Wang}, \bibinfo{person}{Yu Qiao}, \bibinfo{person}{Dahua Lin}, {and} \bibinfo{person}{Bo Dai}.} \bibinfo{year}{2023}\natexlab{}.
\newblock \showarticletitle{Animatediff: Animate your personalized text-to-image diffusion models without specific tuning}.
\newblock \bibinfo{journal}{\emph{arXiv preprint arXiv:2307.04725}} (\bibinfo{year}{2023}).
\newblock


\bibitem[Hertz et~al\mbox{.}(2022)]%
        {p2p}
\bibfield{author}{\bibinfo{person}{Amir Hertz}, \bibinfo{person}{Ron Mokady}, \bibinfo{person}{Jay Tenenbaum}, \bibinfo{person}{Kfir Aberman}, \bibinfo{person}{Yael Pritch}, {and} \bibinfo{person}{Daniel Cohen-Or}.} \bibinfo{year}{2022}\natexlab{}.
\newblock \showarticletitle{Prompt-to-prompt image editing with cross attention control}.
\newblock \bibinfo{journal}{\emph{arXiv preprint arXiv:2208.01626}} (\bibinfo{year}{2022}).
\newblock


\bibitem[Hessel et~al\mbox{.}(2021)]%
        {clipscore}
\bibfield{author}{\bibinfo{person}{Jack Hessel}, \bibinfo{person}{Ari Holtzman}, \bibinfo{person}{Maxwell Forbes}, \bibinfo{person}{Ronan~Le Bras}, {and} \bibinfo{person}{Yejin Choi}.} \bibinfo{year}{2021}\natexlab{}.
\newblock \showarticletitle{Clipscore: A reference-free evaluation metric for image captioning}.
\newblock \bibinfo{journal}{\emph{arXiv preprint arXiv:2104.08718}} (\bibinfo{year}{2021}).
\newblock


\bibitem[Heusel et~al\mbox{.}(2017)]%
        {fid}
\bibfield{author}{\bibinfo{person}{Martin Heusel}, \bibinfo{person}{Hubert Ramsauer}, \bibinfo{person}{Thomas Unterthiner}, \bibinfo{person}{Bernhard Nessler}, {and} \bibinfo{person}{Sepp Hochreiter}.} \bibinfo{year}{2017}\natexlab{}.
\newblock \showarticletitle{Gans trained by a two time-scale update rule converge to a local nash equilibrium}.
\newblock \bibinfo{journal}{\emph{Advances in neural information processing systems}}  \bibinfo{volume}{30} (\bibinfo{year}{2017}).
\newblock


\bibitem[Ho et~al\mbox{.}(2022a)]%
        {imagenvideo}
\bibfield{author}{\bibinfo{person}{Jonathan Ho}, \bibinfo{person}{William Chan}, \bibinfo{person}{Chitwan Saharia}, \bibinfo{person}{Jay Whang}, \bibinfo{person}{Ruiqi Gao}, \bibinfo{person}{Alexey Gritsenko}, \bibinfo{person}{Diederik~P Kingma}, \bibinfo{person}{Ben Poole}, \bibinfo{person}{Mohammad Norouzi}, \bibinfo{person}{David~J Fleet}, {et~al\mbox{.}}} \bibinfo{year}{2022}\natexlab{a}.
\newblock \showarticletitle{Imagen video: High definition video generation with diffusion models}.
\newblock \bibinfo{journal}{\emph{arXiv preprint arXiv:2210.02303}} (\bibinfo{year}{2022}).
\newblock


\bibitem[Ho et~al\mbox{.}(2020)]%
        {ddpm}
\bibfield{author}{\bibinfo{person}{Jonathan Ho}, \bibinfo{person}{Ajay Jain}, {and} \bibinfo{person}{Pieter Abbeel}.} \bibinfo{year}{2020}\natexlab{}.
\newblock \showarticletitle{Denoising diffusion probabilistic models}.
\newblock \bibinfo{journal}{\emph{Advances in Neural Information Processing Systems}}  \bibinfo{volume}{33} (\bibinfo{year}{2020}), \bibinfo{pages}{6840--6851}.
\newblock


\bibitem[Ho and Salimans(2022)]%
        {cfg}
\bibfield{author}{\bibinfo{person}{Jonathan Ho} {and} \bibinfo{person}{Tim Salimans}.} \bibinfo{year}{2022}\natexlab{}.
\newblock \showarticletitle{Classifier-free diffusion guidance}.
\newblock \bibinfo{journal}{\emph{arXiv preprint arXiv:2207.12598}} (\bibinfo{year}{2022}).
\newblock


\bibitem[Ho et~al\mbox{.}(2022b)]%
        {vdm}
\bibfield{author}{\bibinfo{person}{Jonathan Ho}, \bibinfo{person}{Tim Salimans}, \bibinfo{person}{Alexey Gritsenko}, \bibinfo{person}{William Chan}, \bibinfo{person}{Mohammad Norouzi}, {and} \bibinfo{person}{David~J Fleet}.} \bibinfo{year}{2022}\natexlab{b}.
\newblock \showarticletitle{Video diffusion models}.
\newblock \bibinfo{journal}{\emph{arXiv:2204.03458}} (\bibinfo{year}{2022}).
\newblock


\bibitem[Hu et~al\mbox{.}(2021)]%
        {lora}
\bibfield{author}{\bibinfo{person}{Edward~J Hu}, \bibinfo{person}{Yelong Shen}, \bibinfo{person}{Phillip Wallis}, \bibinfo{person}{Zeyuan Allen-Zhu}, \bibinfo{person}{Yuanzhi Li}, \bibinfo{person}{Shean Wang}, \bibinfo{person}{Lu Wang}, {and} \bibinfo{person}{Weizhu Chen}.} \bibinfo{year}{2021}\natexlab{}.
\newblock \showarticletitle{Lora: Low-rank adaptation of large language models}.
\newblock \bibinfo{journal}{\emph{arXiv preprint arXiv:2106.09685}} (\bibinfo{year}{2021}).
\newblock


\bibitem[Hu et~al\mbox{.}(2023)]%
        {animateanyone}
\bibfield{author}{\bibinfo{person}{Li Hu}, \bibinfo{person}{Xin Gao}, \bibinfo{person}{Peng Zhang}, \bibinfo{person}{Ke Sun}, \bibinfo{person}{Bang Zhang}, {and} \bibinfo{person}{Liefeng Bo}.} \bibinfo{year}{2023}\natexlab{}.
\newblock \showarticletitle{Animate Anyone: Consistent and Controllable Image-to-Video Synthesis for Character Animation}.
\newblock \bibinfo{journal}{\emph{arXiv preprint arXiv:2311.17117}} (\bibinfo{year}{2023}).
\newblock


\bibitem[Jain et~al\mbox{.}(2023)]%
        {peekaboo}
\bibfield{author}{\bibinfo{person}{Yash Jain}, \bibinfo{person}{Anshul Nasery}, \bibinfo{person}{Vibhav Vineet}, {and} \bibinfo{person}{Harkirat Behl}.} \bibinfo{year}{2023}\natexlab{}.
\newblock \showarticletitle{PEEKABOO: Interactive Video Generation via Masked-Diffusion}.
\newblock \bibinfo{journal}{\emph{arXiv preprint arXiv:2312.07509}} (\bibinfo{year}{2023}).
\newblock


\bibitem[Jeong et~al\mbox{.}(2023)]%
        {vmc}
\bibfield{author}{\bibinfo{person}{Hyeonho Jeong}, \bibinfo{person}{Geon~Yeong Park}, {and} \bibinfo{person}{Jong~Chul Ye}.} \bibinfo{year}{2023}\natexlab{}.
\newblock \showarticletitle{VMC: Video Motion Customization using Temporal Attention Adaption for Text-to-Video Diffusion Models}.
\newblock \bibinfo{journal}{\emph{arXiv preprint arXiv:2312.00845}} (\bibinfo{year}{2023}).
\newblock


\bibitem[Kasten et~al\mbox{.}(2021)]%
        {lna}
\bibfield{author}{\bibinfo{person}{Yoni Kasten}, \bibinfo{person}{Dolev Ofri}, \bibinfo{person}{Oliver Wang}, {and} \bibinfo{person}{Tali Dekel}.} \bibinfo{year}{2021}\natexlab{}.
\newblock \showarticletitle{Layered neural atlases for consistent video editing}.
\newblock \bibinfo{journal}{\emph{ACM Transactions on Graphics (TOG)}} \bibinfo{volume}{40}, \bibinfo{number}{6} (\bibinfo{year}{2021}), \bibinfo{pages}{1--12}.
\newblock


\bibitem[Kim et~al\mbox{.}(2023)]%
        {densediff}
\bibfield{author}{\bibinfo{person}{Yunji Kim}, \bibinfo{person}{Jiyoung Lee}, \bibinfo{person}{Jin-Hwa Kim}, \bibinfo{person}{Jung-Woo Ha}, {and} \bibinfo{person}{Jun-Yan Zhu}.} \bibinfo{year}{2023}\natexlab{}.
\newblock \showarticletitle{Dense text-to-image generation with attention modulation}. In \bibinfo{booktitle}{\emph{Proceedings of the IEEE/CVF International Conference on Computer Vision}}. \bibinfo{pages}{7701--7711}.
\newblock


\bibitem[Kumari et~al\mbox{.}(2023)]%
        {customdiff}
\bibfield{author}{\bibinfo{person}{Nupur Kumari}, \bibinfo{person}{Bingliang Zhang}, \bibinfo{person}{Richard Zhang}, \bibinfo{person}{Eli Shechtman}, {and} \bibinfo{person}{Jun-Yan Zhu}.} \bibinfo{year}{2023}\natexlab{}.
\newblock \showarticletitle{Multi-concept customization of text-to-image diffusion}. In \bibinfo{booktitle}{\emph{Proceedings of the IEEE/CVF Conference on Computer Vision and Pattern Recognition}}. \bibinfo{pages}{1931--1941}.
\newblock


\bibitem[Li et~al\mbox{.}(2023)]%
        {gligen}
\bibfield{author}{\bibinfo{person}{Yuheng Li}, \bibinfo{person}{Haotian Liu}, \bibinfo{person}{Qingyang Wu}, \bibinfo{person}{Fangzhou Mu}, \bibinfo{person}{Jianwei Yang}, \bibinfo{person}{Jianfeng Gao}, \bibinfo{person}{Chunyuan Li}, {and} \bibinfo{person}{Yong~Jae Lee}.} \bibinfo{year}{2023}\natexlab{}.
\newblock \showarticletitle{Gligen: Open-set grounded text-to-image generation}. In \bibinfo{booktitle}{\emph{Proceedings of the IEEE/CVF Conference on Computer Vision and Pattern Recognition}}. \bibinfo{pages}{22511--22521}.
\newblock


\bibitem[Liu et~al\mbox{.}(2023a)]%
        {groundingdino}
\bibfield{author}{\bibinfo{person}{Shilong Liu}, \bibinfo{person}{Zhaoyang Zeng}, \bibinfo{person}{Tianhe Ren}, \bibinfo{person}{Feng Li}, \bibinfo{person}{Hao Zhang}, \bibinfo{person}{Jie Yang}, \bibinfo{person}{Chunyuan Li}, \bibinfo{person}{Jianwei Yang}, \bibinfo{person}{Hang Su}, \bibinfo{person}{Jun Zhu}, {et~al\mbox{.}}} \bibinfo{year}{2023}\natexlab{a}.
\newblock \showarticletitle{Grounding dino: Marrying dino with grounded pre-training for open-set object detection}.
\newblock \bibinfo{journal}{\emph{arXiv preprint arXiv:2303.05499}} (\bibinfo{year}{2023}).
\newblock


\bibitem[Liu et~al\mbox{.}(2023b)]%
        {videop2p}
\bibfield{author}{\bibinfo{person}{Shaoteng Liu}, \bibinfo{person}{Yuechen Zhang}, \bibinfo{person}{Wenbo Li}, \bibinfo{person}{Zhe Lin}, {and} \bibinfo{person}{Jiaya Jia}.} \bibinfo{year}{2023}\natexlab{b}.
\newblock \showarticletitle{Video-p2p: Video editing with cross-attention control}.
\newblock \bibinfo{journal}{\emph{arXiv preprint arXiv:2303.04761}} (\bibinfo{year}{2023}).
\newblock


\bibitem[Loshchilov and Hutter(2017)]%
        {adamw}
\bibfield{author}{\bibinfo{person}{Ilya Loshchilov} {and} \bibinfo{person}{Frank Hutter}.} \bibinfo{year}{2017}\natexlab{}.
\newblock \showarticletitle{Decoupled weight decay regularization}.
\newblock \bibinfo{journal}{\emph{arXiv preprint arXiv:1711.05101}} (\bibinfo{year}{2017}).
\newblock


\bibitem[Ma et~al\mbox{.}(2023)]%
        {directeddiff}
\bibfield{author}{\bibinfo{person}{Wan-Duo~Kurt Ma}, \bibinfo{person}{JP Lewis}, \bibinfo{person}{W~Bastiaan Kleijn}, {and} \bibinfo{person}{Thomas Leung}.} \bibinfo{year}{2023}\natexlab{}.
\newblock \showarticletitle{Directed diffusion: Direct control of object placement through attention guidance}.
\newblock \bibinfo{journal}{\emph{arXiv preprint arXiv:2302.13153}} (\bibinfo{year}{2023}).
\newblock


\bibitem[Mildenhall et~al\mbox{.}(2021)]%
        {nerf}
\bibfield{author}{\bibinfo{person}{Ben Mildenhall}, \bibinfo{person}{Pratul~P Srinivasan}, \bibinfo{person}{Matthew Tancik}, \bibinfo{person}{Jonathan~T Barron}, \bibinfo{person}{Ravi Ramamoorthi}, {and} \bibinfo{person}{Ren Ng}.} \bibinfo{year}{2021}\natexlab{}.
\newblock \showarticletitle{Nerf: Representing scenes as neural radiance fields for view synthesis}.
\newblock \bibinfo{journal}{\emph{Commun. ACM}} \bibinfo{volume}{65}, \bibinfo{number}{1} (\bibinfo{year}{2021}), \bibinfo{pages}{99--106}.
\newblock


\bibitem[Mokady et~al\mbox{.}(2023)]%
        {nulltxtinv}
\bibfield{author}{\bibinfo{person}{Ron Mokady}, \bibinfo{person}{Amir Hertz}, \bibinfo{person}{Kfir Aberman}, \bibinfo{person}{Yael Pritch}, {and} \bibinfo{person}{Daniel Cohen-Or}.} \bibinfo{year}{2023}\natexlab{}.
\newblock \showarticletitle{Null-text inversion for editing real images using guided diffusion models}. In \bibinfo{booktitle}{\emph{Proceedings of the IEEE/CVF Conference on Computer Vision and Pattern Recognition}}. \bibinfo{pages}{6038--6047}.
\newblock


\bibitem[Mou et~al\mbox{.}(2023)]%
        {t2iadapter}
\bibfield{author}{\bibinfo{person}{Chong Mou}, \bibinfo{person}{Xintao Wang}, \bibinfo{person}{Liangbin Xie}, \bibinfo{person}{Jian Zhang}, \bibinfo{person}{Zhongang Qi}, \bibinfo{person}{Ying Shan}, {and} \bibinfo{person}{Xiaohu Qie}.} \bibinfo{year}{2023}\natexlab{}.
\newblock \showarticletitle{T2i-adapter: Learning adapters to dig out more controllable ability for text-to-image diffusion models}.
\newblock \bibinfo{journal}{\emph{arXiv preprint arXiv:2302.08453}} (\bibinfo{year}{2023}).
\newblock


\bibitem[Ng et~al\mbox{.}(2022)]%
        {animalkingdom}
\bibfield{author}{\bibinfo{person}{Xun~Long Ng}, \bibinfo{person}{Kian~Eng Ong}, \bibinfo{person}{Qichen Zheng}, \bibinfo{person}{Yun Ni}, \bibinfo{person}{Si~Yong Yeo}, {and} \bibinfo{person}{Jun Liu}.} \bibinfo{year}{2022}\natexlab{}.
\newblock \showarticletitle{Animal Kingdom: A Large and Diverse Dataset for Animal Behavior Understanding}. In \bibinfo{booktitle}{\emph{Proceedings of the IEEE/CVF Conference on Computer Vision and Pattern Recognition (CVPR)}}. \bibinfo{pages}{19023--19034}.
\newblock


\bibitem[Ouyang et~al\mbox{.}(2023)]%
        {codef}
\bibfield{author}{\bibinfo{person}{Hao Ouyang}, \bibinfo{person}{Qiuyu Wang}, \bibinfo{person}{Yuxi Xiao}, \bibinfo{person}{Qingyan Bai}, \bibinfo{person}{Juntao Zhang}, \bibinfo{person}{Kecheng Zheng}, \bibinfo{person}{Xiaowei Zhou}, \bibinfo{person}{Qifeng Chen}, {and} \bibinfo{person}{Yujun Shen}.} \bibinfo{year}{2023}\natexlab{}.
\newblock \showarticletitle{Codef: Content deformation fields for temporally consistent video processing}.
\newblock \bibinfo{journal}{\emph{arXiv preprint arXiv:2308.07926}} (\bibinfo{year}{2023}).
\newblock


\bibitem[Qi et~al\mbox{.}(2023)]%
        {fatezero}
\bibfield{author}{\bibinfo{person}{Chenyang Qi}, \bibinfo{person}{Xiaodong Cun}, \bibinfo{person}{Yong Zhang}, \bibinfo{person}{Chenyang Lei}, \bibinfo{person}{Xintao Wang}, \bibinfo{person}{Ying Shan}, {and} \bibinfo{person}{Qifeng Chen}.} \bibinfo{year}{2023}\natexlab{}.
\newblock \showarticletitle{Fatezero: Fusing attentions for zero-shot text-based video editing}.
\newblock \bibinfo{journal}{\emph{arXiv preprint arXiv:2303.09535}} (\bibinfo{year}{2023}).
\newblock


\bibitem[Ramesh et~al\mbox{.}(2022)]%
        {dalle2}
\bibfield{author}{\bibinfo{person}{Aditya Ramesh}, \bibinfo{person}{Prafulla Dhariwal}, \bibinfo{person}{Alex Nichol}, \bibinfo{person}{Casey Chu}, {and} \bibinfo{person}{Mark Chen}.} \bibinfo{year}{2022}\natexlab{}.
\newblock \showarticletitle{Hierarchical text-conditional image generation with clip latents}.
\newblock \bibinfo{journal}{\emph{arXiv preprint arXiv:2204.06125}} (\bibinfo{year}{2022}).
\newblock


\bibitem[Rao et~al\mbox{.}(2020)]%
        {movieshot}
\bibfield{author}{\bibinfo{person}{Anyi Rao}, \bibinfo{person}{Jiaze Wang}, \bibinfo{person}{Linning Xu}, \bibinfo{person}{Xuekun Jiang}, \bibinfo{person}{Qingqiu Huang}, \bibinfo{person}{Bolei Zhou}, {and} \bibinfo{person}{Dahua Lin}.} \bibinfo{year}{2020}\natexlab{}.
\newblock \showarticletitle{A unified framework for shot type classification based on subject centric lens}. In \bibinfo{booktitle}{\emph{Computer Vision--ECCV 2020: 16th European Conference, Glasgow, UK, August 23--28, 2020, Proceedings, Part XI 16}}. Springer, \bibinfo{pages}{17--34}.
\newblock


\bibitem[Rombach et~al\mbox{.}(2022)]%
        {ldm}
\bibfield{author}{\bibinfo{person}{Robin Rombach}, \bibinfo{person}{Andreas Blattmann}, \bibinfo{person}{Dominik Lorenz}, \bibinfo{person}{Patrick Esser}, {and} \bibinfo{person}{Bj{\"o}rn Ommer}.} \bibinfo{year}{2022}\natexlab{}.
\newblock \showarticletitle{High-resolution image synthesis with latent diffusion models}. In \bibinfo{booktitle}{\emph{Proceedings of the IEEE/CVF Conference on Computer Vision and Pattern Recognition}}. \bibinfo{pages}{10684--10695}.
\newblock


\bibitem[Ruiz et~al\mbox{.}(2023)]%
        {db}
\bibfield{author}{\bibinfo{person}{Nataniel Ruiz}, \bibinfo{person}{Yuanzhen Li}, \bibinfo{person}{Varun Jampani}, \bibinfo{person}{Yael Pritch}, \bibinfo{person}{Michael Rubinstein}, {and} \bibinfo{person}{Kfir Aberman}.} \bibinfo{year}{2023}\natexlab{}.
\newblock \showarticletitle{Dreambooth: Fine tuning text-to-image diffusion models for subject-driven generation}. In \bibinfo{booktitle}{\emph{Proceedings of the IEEE/CVF Conference on Computer Vision and Pattern Recognition}}. \bibinfo{pages}{22500--22510}.
\newblock


\bibitem[Saharia et~al\mbox{.}(2022)]%
        {imagen}
\bibfield{author}{\bibinfo{person}{Chitwan Saharia}, \bibinfo{person}{William Chan}, \bibinfo{person}{Saurabh Saxena}, \bibinfo{person}{Lala Li}, \bibinfo{person}{Jay Whang}, \bibinfo{person}{Emily~L Denton}, \bibinfo{person}{Kamyar Ghasemipour}, \bibinfo{person}{Raphael Gontijo~Lopes}, \bibinfo{person}{Burcu Karagol~Ayan}, \bibinfo{person}{Tim Salimans}, {et~al\mbox{.}}} \bibinfo{year}{2022}\natexlab{}.
\newblock \showarticletitle{Photorealistic text-to-image diffusion models with deep language understanding}.
\newblock \bibinfo{journal}{\emph{Advances in Neural Information Processing Systems}}  \bibinfo{volume}{35} (\bibinfo{year}{2022}), \bibinfo{pages}{36479--36494}.
\newblock


\bibitem[Sarukkai et~al\mbox{.}(2024)]%
        {collagediff}
\bibfield{author}{\bibinfo{person}{Vishnu Sarukkai}, \bibinfo{person}{Linden Li}, \bibinfo{person}{Arden Ma}, \bibinfo{person}{Christopher R{\'e}}, {and} \bibinfo{person}{Kayvon Fatahalian}.} \bibinfo{year}{2024}\natexlab{}.
\newblock \showarticletitle{Collage diffusion}. In \bibinfo{booktitle}{\emph{Proceedings of the IEEE/CVF Winter Conference on Applications of Computer Vision}}. \bibinfo{pages}{4208--4217}.
\newblock


\bibitem[Shi et~al\mbox{.}(2023)]%
        {videoflow}
\bibfield{author}{\bibinfo{person}{Xiaoyu Shi}, \bibinfo{person}{Zhaoyang Huang}, \bibinfo{person}{Weikang Bian}, \bibinfo{person}{Dasong Li}, \bibinfo{person}{Manyuan Zhang}, \bibinfo{person}{Ka~Chun Cheung}, \bibinfo{person}{Simon See}, \bibinfo{person}{Hongwei Qin}, \bibinfo{person}{Jifeng Dai}, {and} \bibinfo{person}{Hongsheng Li}.} \bibinfo{year}{2023}\natexlab{}.
\newblock \showarticletitle{Videoflow: Exploiting temporal cues for multi-frame optical flow estimation}.
\newblock \bibinfo{journal}{\emph{arXiv preprint arXiv:2303.08340}} (\bibinfo{year}{2023}).
\newblock


\bibitem[Singer et~al\mbox{.}(2022)]%
        {makeavideo}
\bibfield{author}{\bibinfo{person}{Uriel Singer}, \bibinfo{person}{Adam Polyak}, \bibinfo{person}{Thomas Hayes}, \bibinfo{person}{Xi Yin}, \bibinfo{person}{Jie An}, \bibinfo{person}{Songyang Zhang}, \bibinfo{person}{Qiyuan Hu}, \bibinfo{person}{Harry Yang}, \bibinfo{person}{Oron Ashual}, \bibinfo{person}{Oran Gafni}, {et~al\mbox{.}}} \bibinfo{year}{2022}\natexlab{}.
\newblock \showarticletitle{Make-a-video: Text-to-video generation without text-video data}.
\newblock \bibinfo{journal}{\emph{arXiv preprint arXiv:2209.14792}} (\bibinfo{year}{2022}).
\newblock


\bibitem[Song et~al\mbox{.}(2020)]%
        {ddim}
\bibfield{author}{\bibinfo{person}{Jiaming Song}, \bibinfo{person}{Chenlin Meng}, {and} \bibinfo{person}{Stefano Ermon}.} \bibinfo{year}{2020}\natexlab{}.
\newblock \showarticletitle{Denoising diffusion implicit models}.
\newblock \bibinfo{journal}{\emph{arXiv preprint arXiv:2010.02502}} (\bibinfo{year}{2020}).
\newblock


\bibitem[Tang et~al\mbox{.}(2023)]%
        {dift}
\bibfield{author}{\bibinfo{person}{Luming Tang}, \bibinfo{person}{Menglin Jia}, \bibinfo{person}{Qianqian Wang}, \bibinfo{person}{Cheng~Perng Phoo}, {and} \bibinfo{person}{Bharath Hariharan}.} \bibinfo{year}{2023}\natexlab{}.
\newblock \showarticletitle{Emergent Correspondence from Image Diffusion}.
\newblock \bibinfo{journal}{\emph{arXiv preprint arXiv:2306.03881}} (\bibinfo{year}{2023}).
\newblock


\bibitem[Unterthiner et~al\mbox{.}(2018)]%
        {fvd}
\bibfield{author}{\bibinfo{person}{Thomas Unterthiner}, \bibinfo{person}{Sjoerd Van~Steenkiste}, \bibinfo{person}{Karol Kurach}, \bibinfo{person}{Raphael Marinier}, \bibinfo{person}{Marcin Michalski}, {and} \bibinfo{person}{Sylvain Gelly}.} \bibinfo{year}{2018}\natexlab{}.
\newblock \showarticletitle{Towards accurate generative models of video: A new metric \& challenges}.
\newblock \bibinfo{journal}{\emph{arXiv preprint arXiv:1812.01717}} (\bibinfo{year}{2018}).
\newblock


\bibitem[Wang et~al\mbox{.}(2023a)]%
        {luckydog1}
\bibfield{author}{\bibinfo{person}{Jun Wang}, \bibinfo{person}{Bohan Lei}, \bibinfo{person}{Liya Ding}, \bibinfo{person}{Xiaoyin Xu}, \bibinfo{person}{Xianfeng Gu}, {and} \bibinfo{person}{Min Zhang}.} \bibinfo{year}{2023}\natexlab{a}.
\newblock \showarticletitle{Autoencoder-based conditional optimal transport generative adversarial network for medical image generation}.
\newblock \bibinfo{journal}{\emph{Visual Informatics}} (\bibinfo{year}{2023}).
\newblock


\bibitem[Wang et~al\mbox{.}(2023d)]%
        {modelscope}
\bibfield{author}{\bibinfo{person}{Jiuniu Wang}, \bibinfo{person}{Hangjie Yuan}, \bibinfo{person}{Dayou Chen}, \bibinfo{person}{Yingya Zhang}, \bibinfo{person}{Xiang Wang}, {and} \bibinfo{person}{Shiwei Zhang}.} \bibinfo{year}{2023}\natexlab{d}.
\newblock \showarticletitle{Modelscope text-to-video technical report}.
\newblock \bibinfo{journal}{\emph{arXiv preprint arXiv:2308.06571}} (\bibinfo{year}{2023}).
\newblock


\bibitem[Wang et~al\mbox{.}(2023b)]%
        {disco}
\bibfield{author}{\bibinfo{person}{Tan Wang}, \bibinfo{person}{Linjie Li}, \bibinfo{person}{Kevin Lin}, \bibinfo{person}{Yuanhao Zhai}, \bibinfo{person}{Chung-Ching Lin}, \bibinfo{person}{Zhengyuan Yang}, \bibinfo{person}{Hanwang Zhang}, \bibinfo{person}{Zicheng Liu}, {and} \bibinfo{person}{Lijuan Wang}.} \bibinfo{year}{2023}\natexlab{b}.
\newblock \showarticletitle{DisCo: Disentangled Control for Realistic Human Dance Generation}.
\newblock \bibinfo{journal}{\emph{arXiv preprint arXiv:2307.00040}} (\bibinfo{year}{2023}).
\newblock


\bibitem[Wang et~al\mbox{.}(2023c)]%
        {v2vzero}
\bibfield{author}{\bibinfo{person}{Wen Wang}, \bibinfo{person}{Kangyang Xie}, \bibinfo{person}{Zide Liu}, \bibinfo{person}{Hao Chen}, \bibinfo{person}{Yue Cao}, \bibinfo{person}{Xinlong Wang}, {and} \bibinfo{person}{Chunhua Shen}.} \bibinfo{year}{2023}\natexlab{c}.
\newblock \showarticletitle{Zero-shot video editing using off-the-shelf image diffusion models}.
\newblock \bibinfo{journal}{\emph{arXiv preprint arXiv:2303.17599}} (\bibinfo{year}{2023}).
\newblock


\bibitem[Wang et~al\mbox{.}(2023f)]%
        {videocomposer}
\bibfield{author}{\bibinfo{person}{Xiang Wang}, \bibinfo{person}{Hangjie Yuan}, \bibinfo{person}{Shiwei Zhang}, \bibinfo{person}{Dayou Chen}, \bibinfo{person}{Jiuniu Wang}, \bibinfo{person}{Yingya Zhang}, \bibinfo{person}{Yujun Shen}, \bibinfo{person}{Deli Zhao}, {and} \bibinfo{person}{Jingren Zhou}.} \bibinfo{year}{2023}\natexlab{f}.
\newblock \showarticletitle{VideoComposer: Compositional Video Synthesis with Motion Controllability}. In \bibinfo{booktitle}{\emph{Advances in Neural Information Processing Systems}}. \bibinfo{pages}{7594--7611}.
\newblock


\bibitem[Wang et~al\mbox{.}(2023e)]%
        {motionctrl}
\bibfield{author}{\bibinfo{person}{Zhouxia Wang}, \bibinfo{person}{Ziyang Yuan}, \bibinfo{person}{Xintao Wang}, \bibinfo{person}{Tianshui Chen}, \bibinfo{person}{Menghan Xia}, \bibinfo{person}{Ping Luo}, {and} \bibinfo{person}{Ying Shan}.} \bibinfo{year}{2023}\natexlab{e}.
\newblock \showarticletitle{MotionCtrl: A Unified and Flexible Motion Controller for Video Generation}.
\newblock \bibinfo{journal}{\emph{arXiv preprint arXiv:2312.03641}} (\bibinfo{year}{2023}).
\newblock


\bibitem[Wei et~al\mbox{.}(2023)]%
        {dreamvideo}
\bibfield{author}{\bibinfo{person}{Yujie Wei}, \bibinfo{person}{Shiwei Zhang}, \bibinfo{person}{Zhiwu Qing}, \bibinfo{person}{Hangjie Yuan}, \bibinfo{person}{Zhiheng Liu}, \bibinfo{person}{Yu Liu}, \bibinfo{person}{Yingya Zhang}, \bibinfo{person}{Jingren Zhou}, {and} \bibinfo{person}{Hongming Shan}.} \bibinfo{year}{2023}\natexlab{}.
\newblock \showarticletitle{Dreamvideo: Composing your dream videos with customized subject and motion}.
\newblock \bibinfo{journal}{\emph{arXiv preprint arXiv:2312.04433}} (\bibinfo{year}{2023}).
\newblock


\bibitem[Wu et~al\mbox{.}(2023b)]%
        {tav}
\bibfield{author}{\bibinfo{person}{Jay~Zhangjie Wu}, \bibinfo{person}{Yixiao Ge}, \bibinfo{person}{Xintao Wang}, \bibinfo{person}{Stan~Weixian Lei}, \bibinfo{person}{Yuchao Gu}, \bibinfo{person}{Yufei Shi}, \bibinfo{person}{Wynne Hsu}, \bibinfo{person}{Ying Shan}, \bibinfo{person}{Xiaohu Qie}, {and} \bibinfo{person}{Mike~Zheng Shou}.} \bibinfo{year}{2023}\natexlab{b}.
\newblock \showarticletitle{Tune-a-video: One-shot tuning of image diffusion models for text-to-video generation}. In \bibinfo{booktitle}{\emph{Proceedings of the IEEE/CVF International Conference on Computer Vision}}. \bibinfo{pages}{7623--7633}.
\newblock


\bibitem[Wu et~al\mbox{.}(2023a)]%
        {lamp}
\bibfield{author}{\bibinfo{person}{Ruiqi Wu}, \bibinfo{person}{Liangyu Chen}, \bibinfo{person}{Tong Yang}, \bibinfo{person}{Chunle Guo}, \bibinfo{person}{Chongyi Li}, {and} \bibinfo{person}{Xiangyu Zhang}.} \bibinfo{year}{2023}\natexlab{a}.
\newblock \showarticletitle{Lamp: Learn a motion pattern for few-shot-based video generation}.
\newblock \bibinfo{journal}{\emph{arXiv preprint arXiv:2310.10769}} (\bibinfo{year}{2023}).
\newblock


\bibitem[Xu et~al\mbox{.}(2016)]%
        {msrvtt}
\bibfield{author}{\bibinfo{person}{Jun Xu}, \bibinfo{person}{Tao Mei}, \bibinfo{person}{Ting Yao}, {and} \bibinfo{person}{Yong Rui}.} \bibinfo{year}{2016}\natexlab{}.
\newblock \showarticletitle{Msr-vtt: A large video description dataset for bridging video and language}. In \bibinfo{booktitle}{\emph{Proceedings of the IEEE conference on computer vision and pattern recognition}}. \bibinfo{pages}{5288--5296}.
\newblock


\bibitem[Xu et~al\mbox{.}(2023)]%
        {magicanimate}
\bibfield{author}{\bibinfo{person}{Zhongcong Xu}, \bibinfo{person}{Jianfeng Zhang}, \bibinfo{person}{Jun~Hao Liew}, \bibinfo{person}{Hanshu Yan}, \bibinfo{person}{Jia-Wei Liu}, \bibinfo{person}{Chenxu Zhang}, \bibinfo{person}{Jiashi Feng}, {and} \bibinfo{person}{Mike~Zheng Shou}.} \bibinfo{year}{2023}\natexlab{}.
\newblock \showarticletitle{MagicAnimate: Temporally Consistent Human Image Animation using Diffusion Model}.
\newblock \bibinfo{journal}{\emph{arXiv preprint arXiv:2311.16498}} (\bibinfo{year}{2023}).
\newblock


\bibitem[Yang et~al\mbox{.}(2020)]%
        {glide}
\bibfield{author}{\bibinfo{person}{Han Yang}, \bibinfo{person}{Ruimao Zhang}, \bibinfo{person}{Xiaobao Guo}, \bibinfo{person}{Wei Liu}, \bibinfo{person}{Wangmeng Zuo}, {and} \bibinfo{person}{Ping Luo}.} \bibinfo{year}{2020}\natexlab{}.
\newblock \showarticletitle{Towards photo-realistic virtual try-on by adaptively generating-preserving image content}. In \bibinfo{booktitle}{\emph{Proceedings of the IEEE/CVF conference on computer vision and pattern recognition}}. \bibinfo{pages}{7850--7859}.
\newblock


\bibitem[Yang et~al\mbox{.}(2023a)]%
        {unipaint}
\bibfield{author}{\bibinfo{person}{Shiyuan Yang}, \bibinfo{person}{Xiaodong Chen}, {and} \bibinfo{person}{Jing Liao}.} \bibinfo{year}{2023}\natexlab{a}.
\newblock \showarticletitle{Uni-paint: A unified framework for multimodal image inpainting with pretrained diffusion model}. In \bibinfo{booktitle}{\emph{Proceedings of the 31st ACM International Conference on Multimedia}}. \bibinfo{pages}{3190--3199}.
\newblock


\bibitem[Yang et~al\mbox{.}(2023b)]%
        {renderavideo}
\bibfield{author}{\bibinfo{person}{Shuai Yang}, \bibinfo{person}{Yifan Zhou}, \bibinfo{person}{Ziwei Liu}, {and} \bibinfo{person}{Chen~Change Loy}.} \bibinfo{year}{2023}\natexlab{b}.
\newblock \showarticletitle{Rerender A Video: Zero-Shot Text-Guided Video-to-Video Translation}.
\newblock \bibinfo{journal}{\emph{arXiv preprint arXiv:2306.07954}} (\bibinfo{year}{2023}).
\newblock


\bibitem[Yin et~al\mbox{.}(2023)]%
        {dragnvwa}
\bibfield{author}{\bibinfo{person}{Shengming Yin}, \bibinfo{person}{Chenfei Wu}, \bibinfo{person}{Jian Liang}, \bibinfo{person}{Jie Shi}, \bibinfo{person}{Houqiang Li}, \bibinfo{person}{Gong Ming}, {and} \bibinfo{person}{Nan Duan}.} \bibinfo{year}{2023}\natexlab{}.
\newblock \showarticletitle{Dragnuwa: Fine-grained control in video generation by integrating text, image, and trajectory}.
\newblock \bibinfo{journal}{\emph{arXiv preprint arXiv:2308.08089}} (\bibinfo{year}{2023}).
\newblock


\bibitem[Yuan et~al\mbox{.}(2024)]%
        {luckydog2}
\bibfield{author}{\bibinfo{person}{Liang Yuan}, \bibinfo{person}{Dingkun Yan}, \bibinfo{person}{Suguru Saito}, {and} \bibinfo{person}{Issei Fujishiro}.} \bibinfo{year}{2024}\natexlab{}.
\newblock \showarticletitle{DiffMat: Latent diffusion models for image-guided material generation}.
\newblock \bibinfo{journal}{\emph{Visual Informatics}} (\bibinfo{year}{2024}).
\newblock


\bibitem[Zhang et~al\mbox{.}(2023)]%
        {controlnet}
\bibfield{author}{\bibinfo{person}{Lvmin Zhang}, \bibinfo{person}{Anyi Rao}, {and} \bibinfo{person}{Maneesh Agrawala}.} \bibinfo{year}{2023}\natexlab{}.
\newblock \showarticletitle{Adding conditional control to text-to-image diffusion models}. In \bibinfo{booktitle}{\emph{Proceedings of the IEEE/CVF International Conference on Computer Vision}}. \bibinfo{pages}{3836--3847}.
\newblock


\bibitem[Zhao et~al\mbox{.}(2023b)]%
        {controlvideo}
\bibfield{author}{\bibinfo{person}{Min Zhao}, \bibinfo{person}{Rongzhen Wang}, \bibinfo{person}{Fan Bao}, \bibinfo{person}{Chongxuan Li}, {and} \bibinfo{person}{Jun Zhu}.} \bibinfo{year}{2023}\natexlab{b}.
\newblock \showarticletitle{ControlVideo: Adding Conditional Control for One Shot Text-to-Video Editing}.
\newblock \bibinfo{journal}{\emph{arXiv preprint arXiv:2305.17098}} (\bibinfo{year}{2023}).
\newblock


\bibitem[Zhao et~al\mbox{.}(2023a)]%
        {motiondirector}
\bibfield{author}{\bibinfo{person}{Rui Zhao}, \bibinfo{person}{Yuchao Gu}, \bibinfo{person}{Jay~Zhangjie Wu}, \bibinfo{person}{David~Junhao Zhang}, \bibinfo{person}{Jiawei Liu}, \bibinfo{person}{Weijia Wu}, \bibinfo{person}{Jussi Keppo}, {and} \bibinfo{person}{Mike~Zheng Shou}.} \bibinfo{year}{2023}\natexlab{a}.
\newblock \showarticletitle{Motiondirector: Motion customization of text-to-video diffusion models}.
\newblock \bibinfo{journal}{\emph{arXiv preprint arXiv:2310.08465}} (\bibinfo{year}{2023}).
\newblock


\bibitem[Zhou et~al\mbox{.}(2022)]%
        {magicvideo}
\bibfield{author}{\bibinfo{person}{Daquan Zhou}, \bibinfo{person}{Weimin Wang}, \bibinfo{person}{Hanshu Yan}, \bibinfo{person}{Weiwei Lv}, \bibinfo{person}{Yizhe Zhu}, {and} \bibinfo{person}{Jiashi Feng}.} \bibinfo{year}{2022}\natexlab{}.
\newblock \showarticletitle{Magicvideo: Efficient video generation with latent diffusion models}.
\newblock \bibinfo{journal}{\emph{arXiv preprint arXiv:2211.11018}} (\bibinfo{year}{2022}).
\newblock


\end{thebibliography}
\clearpage


\appendix
{\Huge \noindent\textbf{Appendix}}

\section{Additional Implementation Details} \label{sup.impl}

\subsection{Camera Augmentation Details} \label{sup.impl.cam_aug}

Extracting camera movement parameters from real-world videos are computationally intensive, often requiring the cumbersome process of separating object motion from camera movement. To bypass these challenges, we propose a method of camera augmentation that simulates camera movement by algorithmically manipulating a stationary camera's footage. In brief, the camera augmentation is implemented by altering the calculated cropping window across the video sequence captured by a stationary camera, thereby simulating the effect of camera movement in a computationally efficient manner. The detailed pseudo-code of this process is illustrated in Figure~\ref{fig.sup.cam_aug}.

\definecolor{softgreen}{RGB}{0, 153, 0} 
\newcommand{\green}[1]{\textcolor{softgreen}{#1}}

\begin{figure*}
\begin{Verbatim}[commandchars=\\\{\}]
Function aug_with_cam_motion(src_video, cx, cy, cz, h, w):
   
    \green{# \textbf{Parameters}:}
    \green{# src_video: Source video, a tensor with the size of [frames (f), 3, src_height, src_width]}
    \green{# cx: Horizontal translation ratio (-1 to 1)}
    \green{# cy: Vertical translation ratio (-1 to 1)}
    \green{# cz: Zoom ratio (0.5 to 2)}
    \green{# h: Height of the augmented video}
    \green{# w: Width of the augmented video}
    \green{# \textbf{Returns}: Augmented video, a tensor with the size of [f, 3, h, w]}


    \green{# Get source frame number, width and height from src_video}
    f, src_h, src_w = src_video.shape[0], src_video.shape[2], src_video.shape[3]

    \green{# Initialize camera boxes for frame cropping}
    cam_boxes = zeros(f, 4) \green{# f frames, 4: [x1,y1,x2,y2]}

    \green{# Calculate dynamic cropping relative coordinates for each frame}
    \green{# The first frame coordinates is the reference, which is always [0,0,1,1].}
    cam_boxes[:, 0] = linspace(0, cx + (1 - 1/cz) / 2, f)  \green{# x1, top-left x}
    cam_boxes[:, 1] = linspace(0, cy + (1 - 1/cz) / 2, f)  \green{# y1, top-left y}
    cam_boxes[:, 2] = linspace(1, cx + (1 + 1/cz) / 2, f)  \green{# x2, bottom-right x}
    cam_boxes[:, 3] = linspace(1, cy + (1 + 1/cz) / 2, f)  \green{# y2, bottom-right y}

    \green{# Compute the minimum and maximum relative coordinates}
    min_x = min(cam_boxes[:, 0::2])
    max_x = max(cam_boxes[:, 0::2])
    min_y = min(cam_boxes[:, 1::2])
    max_y = max(cam_boxes[:, 1::2])

    \green{# Normalize the camera boxes}
    normalized_boxes = zeros_like(cam_boxes)
    normalized_boxes[:, 0::2] = (cam_boxes[:, 0::2] - min_x) / (max_x - min_x)
    normalized_boxes[:, 1::2] = (cam_boxes[:, 1::2] - min_y) / (max_y - min_y)

    \green{# Initialize a tensor for the new frames}
    augmented_frames = zeros(f, 3, h, w)

    \green{# Process each frame}
    for i in range(f):
        \green{# Calculate the actual cropping coordinates}
        x1, y1, x2, y2 = normalized_boxes[i] * tensor([src_w, src_h, src_w, src_h])
        
        \green{# Crop the frame according to the coordinates}
        crop = src_video[i][:, int(y1):int(y2), int(x1):int(x2)]
        
        \green{# Resize the cropped frame and store it}
        augmented_frames[i] = interpolate(crop, size=(h, w), mode='bilinear')

    return augmented_frames
\end{Verbatim}
\caption{Pseudo-code for the camera augmentation function.} \label{fig.sup.cam_aug}
\end{figure*}

\begin{figure*}[t]
\begin{center}
   \includegraphics[width=\linewidth]{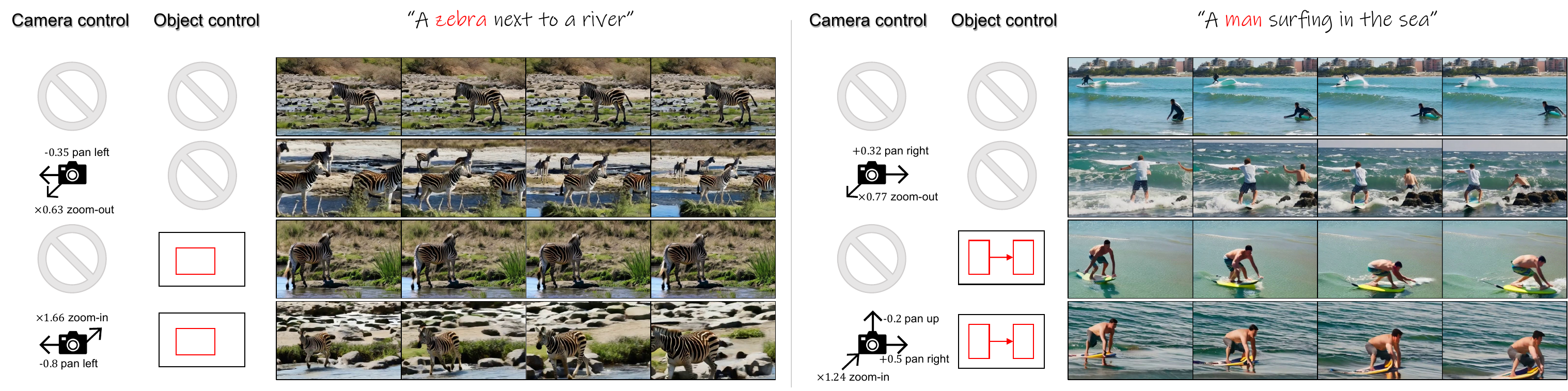}
\end{center}
   \caption{Qualitative comparison of generated videos using the same prompt but different controls. 1st row: base model, \ie, no control; 2nd row: camera control only; 3rd row: object control only; 4th row: camera + object control. Adding control introduces more dynamic content without noticeable quality  degradation.} 
\label{fig.abl.addcontrol}
\end{figure*}

\subsection{Training Details for Camera Control}  \label{sup.impl.cam_train}

\paragraph{Camera movement parameters sampling}
During the training, we adopt the following sampling scheme for camera movement parameters $ \mathbf{c}_\mathrm{cam} =[c_x, c_y, c_z]$:

\begin{align*}
    c_x  &\sim \begin{cases} 
                0, & \text{with probability } \frac{1}{3}, \\
                \text{Uniform}(-1, 1), & \text{with probability } \frac{2}{3},
              \end{cases} \\
    c_y  &\sim \begin{cases} 
                0, & \text{with probability } \frac{1}{3}, \\
                \text{Uniform}(-1, 1), & \text{with probability } \frac{2}{3},
              \end{cases} \\
    c_z &\sim \begin{cases}
                1, & \text{with probability } \frac{1}{3}, \\
                2^\omega, & \text{with probability } \frac{2}{3}, \text{ where } \omega \sim \text{Uniform}(-1, 1).
             \end{cases}
\end{align*}
Note that each component is sampled independently.

\paragraph{Training scheme}
We adopt pretrained Zeroscope T2V model \cite{modelscope} as our base model. To facilitate camera movement learning while retain the pretrained state, only the newly added layers are trainable, which include camera embedder and camera module. To speed up the training, we use a coarse-to-fine strategy: we first train on videos of size $256\times 256\times 8$ (height × width × frames) for 100k iterations, then we resume training on videos of size $320\times 512\times 16$ and $320\times 512\times 24$ for 50k iterations each. The training is performed using a DDPM noise scheduler \cite{ddpm} with timestep $t$ uniformly sampled from $[400,1000]$, such preference to higher timesteps helps to prevents overfitting to low-level details, which are deemed non-essential for understanding temporal transitions. We employ an AdamW optimizer \cite{adamw} with a learning rate of 5e-5 and a batch size of 8 on 8 NVIDIA Tesla V100 GPUs.

\subsection{Inference Details for Camera Control.} \label{sup.impl.cam_inference}
In the text-to-image sampling process, classifier-free guidance \cite{cfg} is widely used to facilitate the text response in generated images, where the predicted noise is extrapolated from the unconditional branch (which uses null-text $\emptyset_\mathrm{txt}$) towards the conditional branch (which uses normal prompt $\mathbf{c}_\mathrm{txt}$). We additionally propose a similar technique to enhance the camera control capability of our model. Specifically, our conditional branch uses desired camera parameters $\mathbf{c}_\mathrm{cam}=[c_x,c_y,c_z]$, while the unconditional branch uses a static camera status $\emptyset_\mathrm{cam}=[0,0,1]$ (\ie, no panning or zooming). The predicted noise $\hat{\epsilon}_{\theta}$ at each sampling step is calculated as:

\begin{equation} \label{eq.cam_cfg}
\begin{split}
& \hat{\epsilon}_{\theta}\left(\mathbf{z}_t,\mathbf{c}_\mathrm{cam},\mathbf{c}_\mathrm{txt},t\right) =  \ \epsilon_{\theta}\left(\mathbf{z}_t, \emptyset_\mathrm{cam},\emptyset_\mathrm{txt}, t \right) \\
& + s \left(\epsilon_{\theta}\left(\mathbf{z}_t,\mathbf{c}_\mathrm{cam},\mathbf{c}_\mathrm{txt},t\right) - \epsilon_{\theta}\left(\mathbf{z}_t, \emptyset_\mathrm{cam},\emptyset_\mathrm{txt}, t \right)\right),
\end{split}
\end{equation}
where $s$ is the guidance scale. On the other hand, unlike text conditioning, which is applied throughout the sampling process, we found that applying the camera conditioning in only a few initial steps is sufficient for controlling camera movement, as the general temporal transitions is already determined in early stages. Formally, during the inference, we bypass the camera module when $t$ is less than a certain threshold, which we refer to as the camera control cut-off timestep, we empirically set this value to $0.85T$.

\section{Additional Ablation Studies} \label{sup.abl}
We conduct additional ablation studies to validate the settings of our method.

\paragraph{Which layers for attention amplification?}
To determine which layers to apply the attention amplification, we divide the U-Net into three parts: encoder (E), middle layer (M), and decoder (D). We applied attention amplification to various combinations of these three and assessed their impact on the CLIP-sim, mIOU and AP50 scores. The results are presented in Table \ref{tab.abl.attn_layer}. 
We observed that applying attention amplification to either the encoder or the decoder significantly enhances object responsiveness, as evidenced by higher values across all metrics. Controllability is further strengthened when attention amplification is applied to both components. The middle layer has a comparatively smaller influence, incorporating middle layer does not bring noticeable statistic change. Consequently, we apply attention amplification across all layers.

\begin{table}
\small
\centering
\caption{Assessment of attention amplification on different parts of UNet.}
\label{tab.abl.attn_layer}
\begin{tabular}{@{}l|ccccccc@{}}
\toprule
 & E & M & D & E+M & M+D & E+D & E+M+D \\   
\midrule
CLIP-sim $\uparrow$ & 26.20 & 25.93 & 26.75 & 26.35 & 26.74 & 27.62 & \textbf{27.63} \\
\R{mIOU(\%) $\uparrow$}   & \R{30.90}   & \R{14.28} & \R{29.25} & \R{31.71}  & \R{28.98} & \R{\textbf{49.06}} &  \R{47.83} \\
\R{AP50(\%) $\uparrow$}   & \R{5.50}   & \R{0.27} & \R{5.75}  & \R{6.61}   & \R{6.13}   & \R{30.04}   &\R{\textbf{31.33}} \\

\bottomrule
\end{tabular}
\end{table}

\begin{figure}[!htpb]
\begin{center}
   \includegraphics[width=\linewidth]{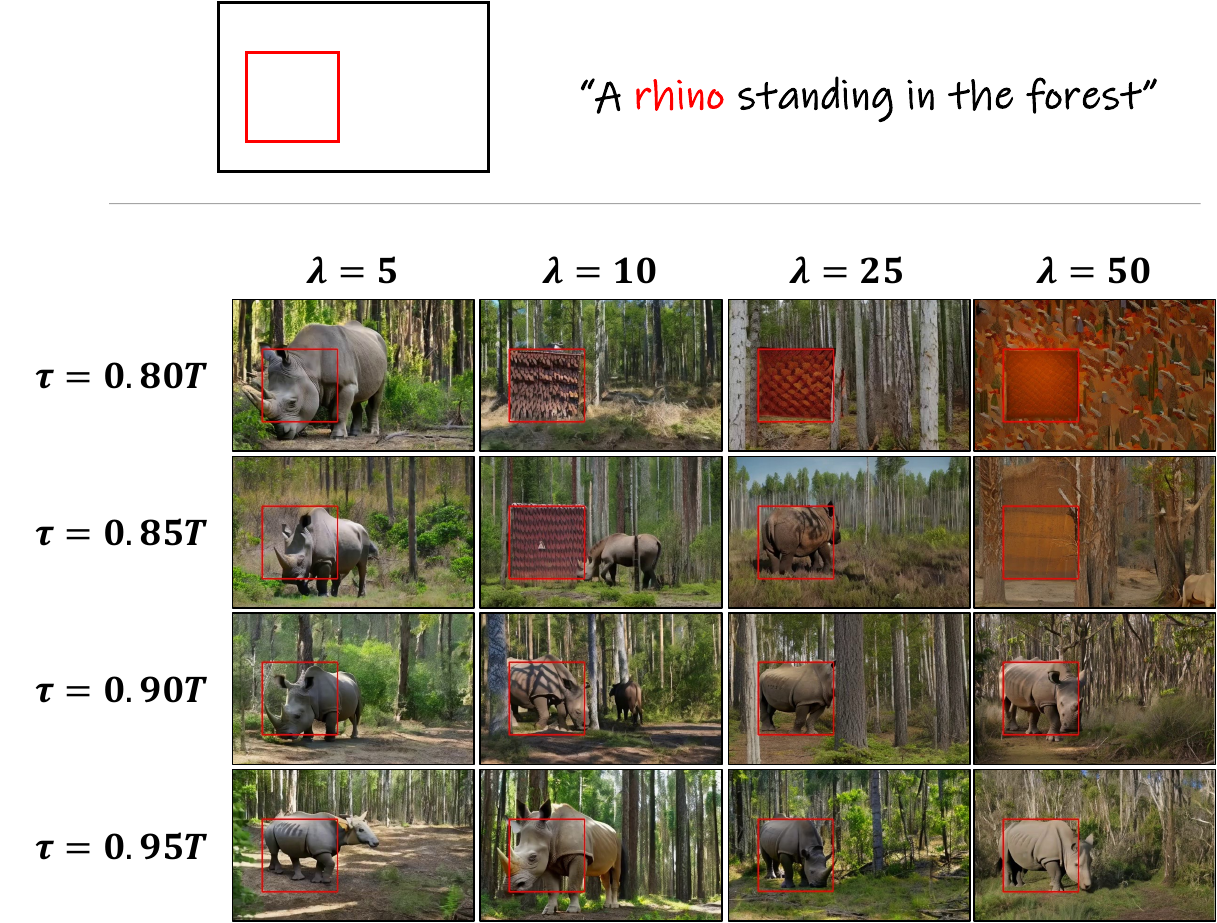}
\end{center}
   \caption{Effect of attention amplification strength $\lambda$ and cut-off timestep $\tau$ (only the first frame is showed).}
\label{fig.abl.attn_param}
\end{figure}

\paragraph{Attention amplification hyper-parameters}
In attention amplification, the strength $\lambda$ and cut-off timestep $\tau$ are two hyper-parameters. Generally, lower $\tau$ and higher $\lambda$ will increase the strength of attention amplification. To determine a proper choice of hyper-parameters,  we conduct tests with different combinations of $\lambda$ and  $\tau$.
Visual examples are provided in Figure \ref{fig.abl.attn_param}. We observed that object responses are more sensitive to the value of $\tau$ than to $\lambda$.  As illustrated in the 1st and 2nd rows, over-responsiveness in box regions typically occurs for $\tau < 0.9$. This is because the early sampling stage in the diffusion model plays a significant role in determining the coarse layout of the output image or video; thus, applying amplification for an extended duration results in over-responsiveness in the box region. \R{We also report CLIP-sim and mIOU metrics in Table \ref{tab.abl.attn_param}, as can be seen, setting $\tau > 0.9$ results in better semantic quality, as evidenced by higher CLIP-sim scores and the visual results. On the other hand, setting $\lambda \geq 10$ generally yields higher mIOU values. It is important to note that while a higher mIOU indicates better object grounding ability, it does not necessarily equate to better object quality.}
In summary, we empirically determine that $\tau \in [0.9T, 0.95T]$ and $\lambda \in [10,25]$ are generally appropriate for most cases.

\begin{table}[htbp]
\centering
\caption{CLIP-sim and mIOU metrics tested on different attention amplification hyper-parameters.}
\label{tab.abl.attn_param}
\footnotesize
\setlength{\tabcolsep}{0.6mm}
\begin{tabular}{@{}l|cc|cc|cc|cc@{}}
\toprule
      & \multicolumn{2}{c|}{$\lambda=5$} & \multicolumn{2}{c|}{$\lambda=10$} & \multicolumn{2}{c|}{$\lambda=25$} & \multicolumn{2}{c}{$\lambda=50$} \\ \cmidrule(l){2-9} 
      & CLIP-sim    & \R{mIOU}     & CLIP-sim     & \R{mIOU}     & CLIP-sim     & \R{mIOU}     & CLIP-sim    & \R{mIOU}     \\ \midrule
$\tau=0.80T$  & 26.81       & \R{27.99} & 24.91        & \R{47.71}  & 24.65        & \R{\textbf{49.76}}  & 24.83       & \R{47.83}   \\
$\tau=0.85T$ & 26.85       & \R{25.72} & 25.31        & \R{42.52}  & 25.03        & \R{47.69}  & 25.29       & \R{47.75}   \\
$\tau=0.90T$  & 26.78       & \R{26.74} & 26.15        & \R{40.83}  & 26.60        & \R{45.50}  & 26.18       & \R{44.08}   \\
$\tau=0.95T$ & 26.48       & \R{21.82} & 27.49        & \R{41.61}  & \textbf{27.63} & \R{47.83}  & 27.32       & \R{43.92}  \\ \bottomrule
\end{tabular}

\end{table}

\paragraph{Effect of adding control on quality.}
To evaluate the impact of incorporating camera/object control on the video quality, we calculate the FVD and FID-vid score under four different settings: (1) Base: no control involved, \ie, the vanilla model; (2) Cam: only camera movement control is involved (with random camera parameters); (3) Obj: only object control is involved; and (4) Cam+Obj: both camera and object control are enabled. The quality metrics are presented in Table~\ref{tab.abl.addcontrol}. The statistic shows that adding control may have a minor influence but not so significant, as the metrics are approximately in the same level with minor fluctuations. We also present visual examples in Figure~\ref{fig.abl.addcontrol}. As can be seen, adding control does not result in a noticeable degradation of quality; on the contrary, it introduces more dynamic content into the generated videos compared to the base model.

\begin{table}
\centering
\caption{Quantitative evaluation for camera/object control on video quality.}
\label{tab.abl.addcontrol}
\begin{tabular}{@{}l|cccc@{}}
\toprule
 & Base & Cam & Obj & Cam+Obj \\ 
\midrule
FID-vid $\downarrow$ & \textbf{41.12} & 44.95 & 43.55 & 41.20 \\
FVD $\downarrow$ & \textbf{1104.36} & 1204.55 & 1300.86 & 1280.88 \\
\bottomrule
\end{tabular}
\end{table}

\section{Additional Results} \label{sup.results}
\begin{figure*}[t]
\begin{center}
   \includegraphics[width=\linewidth]{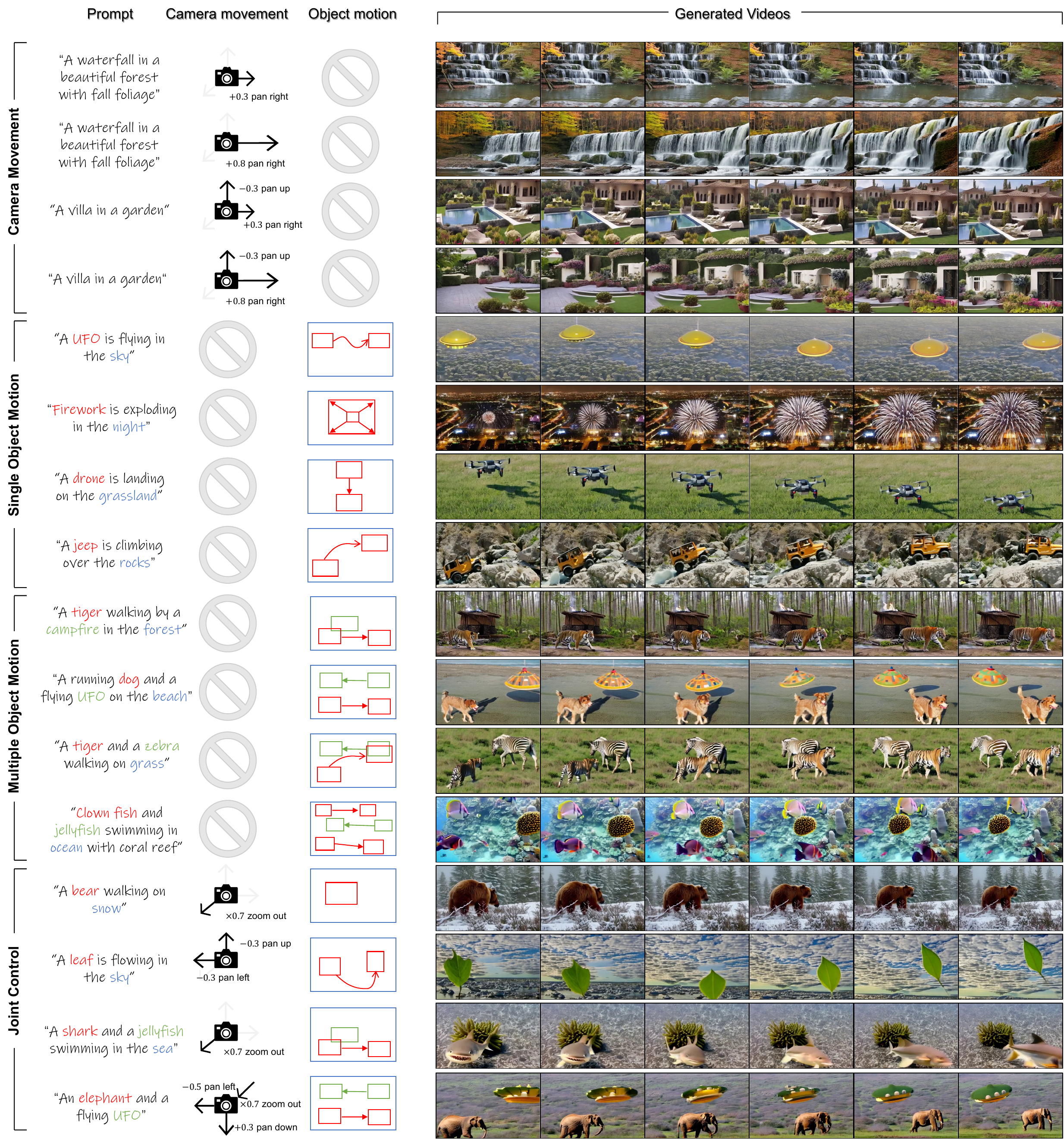}
\end{center}
\caption{Additional results of camera movement control and object motion control.}
\label{fig.extra}
\end{figure*}

We show additional results in Figure~\ref{fig.extra}. Please refer to our project page for dynamic results.

\end{document}